\documentclass[a4paper,fleqn]{cas-dc}

\usepackage[numbers]{natbib}

\usepackage{color,array,amsthm}
\usepackage{graphicx}
\usepackage{subcaption}
\usepackage[ruled, noend]{algorithm2e} 
\usepackage{amsmath}
\usepackage{amssymb}
\usepackage{booktabs}
\usepackage{multirow}
\usepackage{textcomp}
\usepackage{csquotes}
\usepackage{makecell}
\usepackage{float}

\def\tsc#1{\csdef{#1}{\textsc{\lowercase{#1}}\xspace}}
\tsc{WGM}
\tsc{QE}
\tsc{EP}
\tsc{PMS}
\tsc{BEC}
\tsc{DE}

\begin{document}
\let\WriteBookmarks\relax
\def\floatpagepagefraction{1}
\def\textpagefraction{.001}
\shorttitle{TCBiRRT: Rapid Motion Planning for Tightly Coupled Dual-arm Space Manipulator}
\shortauthors{J. Zhang et~al.}

\title [mode = title]{TCBiRRT: Rapid Motion Planning for Tightly Coupled Dual-arm Space Manipulator Using Task-space Random Expansion}



\author[1]{Jiawei Zhang}
\credit{Conceptualization of this study, Methodology, Software, Writing - Original draft preparation}

\author[1]{Xinhao Miao}
\credit{Writing - Original draft preparation}

\author[1]{Jifeng Guo}

\credit{Conceptualization of this study}

\author[1]{Qinghua Li}

\credit{Conceptualization of this study}

\author[1]{Chengchao Bai}
\ead{baichengchao@hit.edu.cn}
\cormark[1]
\credit{Conceptualization of this study}

\affiliation[1]{organization={Harbin Institute of Technology},
	addressline={Xidazhi Street 92}, 
	postcode={150001}, 
	city={Harbin},
	country={China}}
	
\cortext[cor1]{Corresponding author}

\begin{abstract}
Planning the motion path for a tightly coupled dual-arm space manipulator under closed-chain constraints is a fundamental yet challenging problem in on-orbit assembly of large-scale space structures. The closed-chain constraints significantly reduce the feasible configuration space, making it difficult for existing planners to efficiently generate collision-free motions, especially in cluttered environments. To address this issue, this paper proposes a task-space constrained bidirectional rapidly-exploring random tree algorithm, termed TCBiRRT. Unlike conventional methods that operate in the high-dimensional configuration space, the proposed approach performs random sampling and node expansion directly in the task space defined by the manipulated object pose. A task-space node expansion strategy is developed to generate candidate object motions, which are then mapped to continuous joint paths using a path inverse kinematics algorithm. The method is further integrated with a bidirectional RRT framework and a regrasp mechanism to efficiently connect two random trees. Extensive simulations are conducted in representative on-orbit assembly scenarios with varying levels of environmental complexity. The results demonstrate that TCBiRRT achieves significantly higher success rates and orders-of-magnitude improvements in planning time compared to state-of-the-art planners. The proposed method provides an efficient and robust solution for motion planning of tightly coupled dual-arm space manipulators.
\end{abstract}


%
%


\maketitle

\section{Introduction}

The on-orbit assembly of ultra-large spacecraft, such as space solar power stations, large space telescopes, and deep-space transfer hubs, has attracted increasing attention from major space agencies \cite{c1}. These spacecrafts are typically composed of hundreds of modules, including supporting trusses, antenna arrays, battery panels, rotary joints, etc \cite{c2}. The extreme orbital environment makes the assembly tasks rely heavily on the autonomous operation capability of space manipulators.

\begin{figure}
	\centering
	\includegraphics[width=0.49\textwidth]{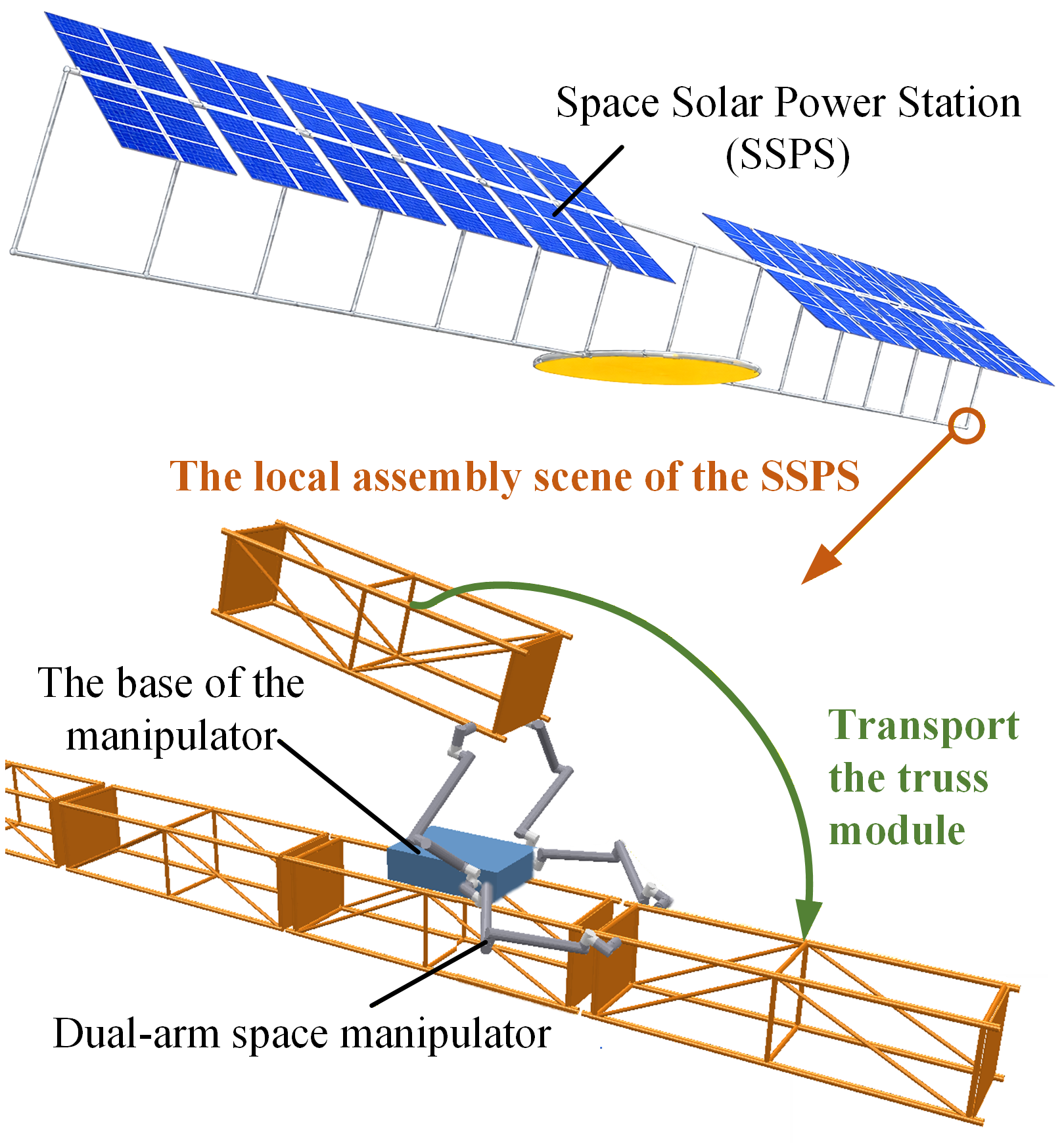} %
	\caption{The tightly coupled dual-arm space manipulator transports the module close to the assembly interface.}
	\label{fig1}
\end{figure}

Compared with a single-arm space manipulator, tightly coupled dual-arm space manipulator with closed-chain constraints \cite{c3}\cite{c4} provide superior load capacity, structural stiffness, and operational precision. These advantages make them a key technology for high-precision assembly of large modules. The assembly task of the modules can be divided into two stages: safe transportation phase \cite{c3} and docking phase \cite{c4}. In the transportation stage, the module must be moved close to the assembly interface, as shown in Fig .\ref{fig1}. One of the core challenges in this phase is to plan a collision-free motion path for the dual-arm manipulator under closed-chain constraints. In particular, the efficiency of motion planning directly affects the overall assembly time, making rapid planning essential for practical deployment.

Existing approaches for motion planning under closed-chain constraints can be broadly categorized into sampling-based methods \cite{c5}, optimization-based methods \cite{c6}, and neural motion planning methods \cite{c3}. Sampling-based methods provide probabilistic completeness but often suffer from low efficiency in cluttered environments due to the difficulty of sampling valid configurations on the constraint manifold. Optimization-based methods can generate smooth trajectories efficiently, but they rely heavily on the quality of the initial trajectory and lack completeness. Neural motion planning methods achieve rapid planning, but require extensive offline training and exhibit limited generalization when task conditions change. These limitations indicate that simultaneously achieving planning efficiency, completeness, and robustness under closed-chain constraints remains an open challenge.

To address these challenges, this paper proposes a Task-space Constrained Bidirectional Rapidly-exploring Random Tree method, referred to as TCBiRRT. Instead of performing exploration in the high-dimensional configuration space, the proposed method expands the nodes in the task space. By integrating a path inverse kinematics solver, continuous joint trajectories can be generated from task-space paths while satisfying closed-chain constraints. The main contributions of this paper are as follows:

\begin{enumerate}
	\def\labelenumi{\arabic{enumi})}
	\item
	A node extension method is proposed for random trees in the task space, which transforms the complex constrained motion planning problem into a lower-dimensional planning problem in the task space, significantly improving the efficiency of node extension.
	
	\item
	The proposed task-space node expansion method is integrated with a Bidirectional Rapidly-exploring Random Trees (BiRRT) framework to develop the TCBiRRT algorithm, and a regrasp mechanism is utilized to quickly find a feasible paths.
	
	\item
	Extensive simulation tests of the proposed TCBiRRT algorithm were conducted in representative on-orbit assembly scenarios. The results demonstrate that the proposed method achieves significantly faster planning speed compared with state-of-the-art motion planners.
\end{enumerate}

\section{Related Work}

\subsection{Sampling-Based Constrained Motion Planning}

When there are closed-chain constraints, all feasible configurations form a lower-dimensional constraint manifold embedded in the configuration space. Since the probability of sampling directly on this manifold is zero, classical sampling-based motion planning algorithms such as Probabilistic Roadmap (PRM) \cite{c7} and RRT \cite{c8} cannot be directly applied. To address this issue, sampling-based constrained motion planning methods modify the sampling strategy or local planner to handle constraints.

Constraint relaxation methods \cite{c9} \cite{c10} allow temporary violations of constraints within a small predefined tolerance, which effectively enlarges the feasible region in the configuration space. Although these methods require minimal modification to existing planners, their sampling efficiency remains low because the allowable violation is typically very small.

Projection-based methods first generate samples in the configuration space and then project them onto the constraint manifold. Projection-based methods can be divided into two categories: inverse-kinematics-based methods \cite{c11} \cite{c12} \cite{c13} and numerical methods \cite{c5} \cite{c14} \cite{c15}. Inverse-kinematics-based projection methods split the closed-chain system into active and passive chains, but the existence of valid solutions is not guaranteed. Keunwoo et al. \cite{c15} extended the PRM algorithm to the motion planning problem with closed-chain constraints, obtains feasible nodes by randomly sampling the object pose, followed by computing the inverse kinematics solution of the multi manipulators. Numerical methods, such as Newton-Raphson projection method \cite{c5} \cite{c16} \cite{c17}, improve efficiency by iteratively converging to the constraint manifold using the pseudoinverse of the constraint Jacobian matrix.

To further improve efficiency, the null space of the constraint Jacobian matrix serves as the tangent space of the constraint manifold, and the tangent-space-based methods approximate the constraint manifold locally using the tangent space and perform sampling in the tangent space \cite{c18}.  Atlas-based methods \cite{c19} extend this idea by incrementally constructing a global approximation of the manifold, reducing repeated computations of the tangent space and improving planning efficiency.

Data-driven approaches have recently been introduced to accelerate sampling. The precomputed roadmap \cite{c20} method approximates the constraint manifold offline, but lack flexibility when constraints change. Learning-based samplers \cite{c21} and latent space representations \cite{c22} \cite{c23} \cite{c24} improve sampling efficiency, yet their performance depends heavily on training data quality and may degrade in cluttered environments.

Although sampling-based methods provide probabilistic completeness, their performance degrades significantly in environments with dense obstacles due to the difficulty of generating valid samples on the constraint manifold.

\subsection{Optimization-Based Constrained Motion Planning}

Optimization-based methods formulate motion planning as a trajectory optimization problem, which typically take the initial trajectory as input and optimize the smoothness, length, obstacle avoidance of the trajectory. A representative method is the Covariant Hamiltonian Optimization for Motion Planning (CHOMP), which has been used for constrained motion planning \cite{c25}, and a multigrid CHOMP was proposed to speed up convergence by alternately performing optimization and upsampling \cite{c26}. However, collision constraints between the manipulators and obstacles were not explicitly considered during the trajectory optimization process of CHOMP. Szynkiewicz et al. \cite{c6} formulated the closed-chain constrained motion planning problem as a quasi-static nonlinear programming problem with equality and inequality constraints and solved it numerically, which also did not consider collision constraints. Sequential convex optimization methods, such as TrajOpt \cite{c27}, reformulate the planning problem into a series of convex subproblems and incorporate collision avoidance using penalty-based formulations. TrajOpt can explicitly consider continuous-time collision safety, which improves robustness in cluttered environments. Völz et al. \cite{c28} formulated the problem as dynamic programming problem, reporting higher success rates and better scalability than CHOMP and TrajOpt. The optimization-based methods require the trajectory to be discretized, and the discretization can cause deviations from the constraint manifold, making it difficult for the controller to track the planned path. Bordalba et al. \cite{c29} introduced the manifold geometric integration techniques to address this problem.

Optimization-based planners produce high-quality trajectories efficiently but rely on the quality of the initial trajectory and lack completeness, their success rate deteriorates in clutter environments.

\subsection{Neural Motion Planning}

Neural motion planning (NMP) methods \cite{c30} \cite{c31} leverage neural networks to learn motion planning policies directly from data. Unlike traditional planners, which rely on explicit modeling and search processes, NMP methods approximate the mapping from states to actions or trajectories through offline training. This paradigm enables fast online plan and makes NMP particularly attractive for high-dimensional planning problems. For systems with closed-chain constraints, Zhang et al. \cite{c3} proposed a reinforcement-learning-based NMP algorithm. Their method uses an object-centric action space and an object-centric state transition process to handle closed-chain constraints, and introduces a reward function conditioned on goal connectivity to reduce training difficulty.

Although NMP methods can achieve impressive performance after training, they require substantial amounts of training data. In addition, their generalization capability remains limited when the task environment changes significantly. As a result, these methods are more suitable for scenarios with repetitive tasks and relatively static environments.

\section{Problem Statement}

Suppose there are two space manipulators with a common base, and this base is attached to a large spacecraft. The coordinate system fixed to the base is denoted as ${\sum _b}$. The coordinate system fixed at the end of the space manipulator is denoted as ${\sum _i}(i = 1,2)$. The position vector, rotation matrix and homogeneous transformation matrix of the end coordinate system of manipulator $i$ in the ${\sum _b}$ are represented by ${\boldsymbol{p}_i}({\boldsymbol{q}_i}) \in {\mathbb{R}^3}$, ${\boldsymbol{R}_i}({\boldsymbol{q}_i}) \in SO(3)$ and ${\boldsymbol{T}_i}({\boldsymbol{q}_i}) \in SE(3)$ respectively, which can be calculated using the forward kinematics model of the manipulator $i$. Assume that the degrees of freedom of each manipulator is $n$. The joint angle vector of manipulator $i$ is represented as ${\boldsymbol{q}_i} \in {\mathbb{R}^n}$, and $\boldsymbol{q}{\rm{ = [}}\boldsymbol{q}_1^{\rm{T}}{\rm{,}}\boldsymbol{q}_2^{\rm{T}}{{\rm{]}}^{\rm{T}}} \in {\mathbb{R}^{2n}}$ represents the joint angle vectors of all the manipulators.

\begin{figure}
	\centering
	\setlength{\abovecaptionskip}{0.cm}
	\includegraphics[width=0.49\textwidth]{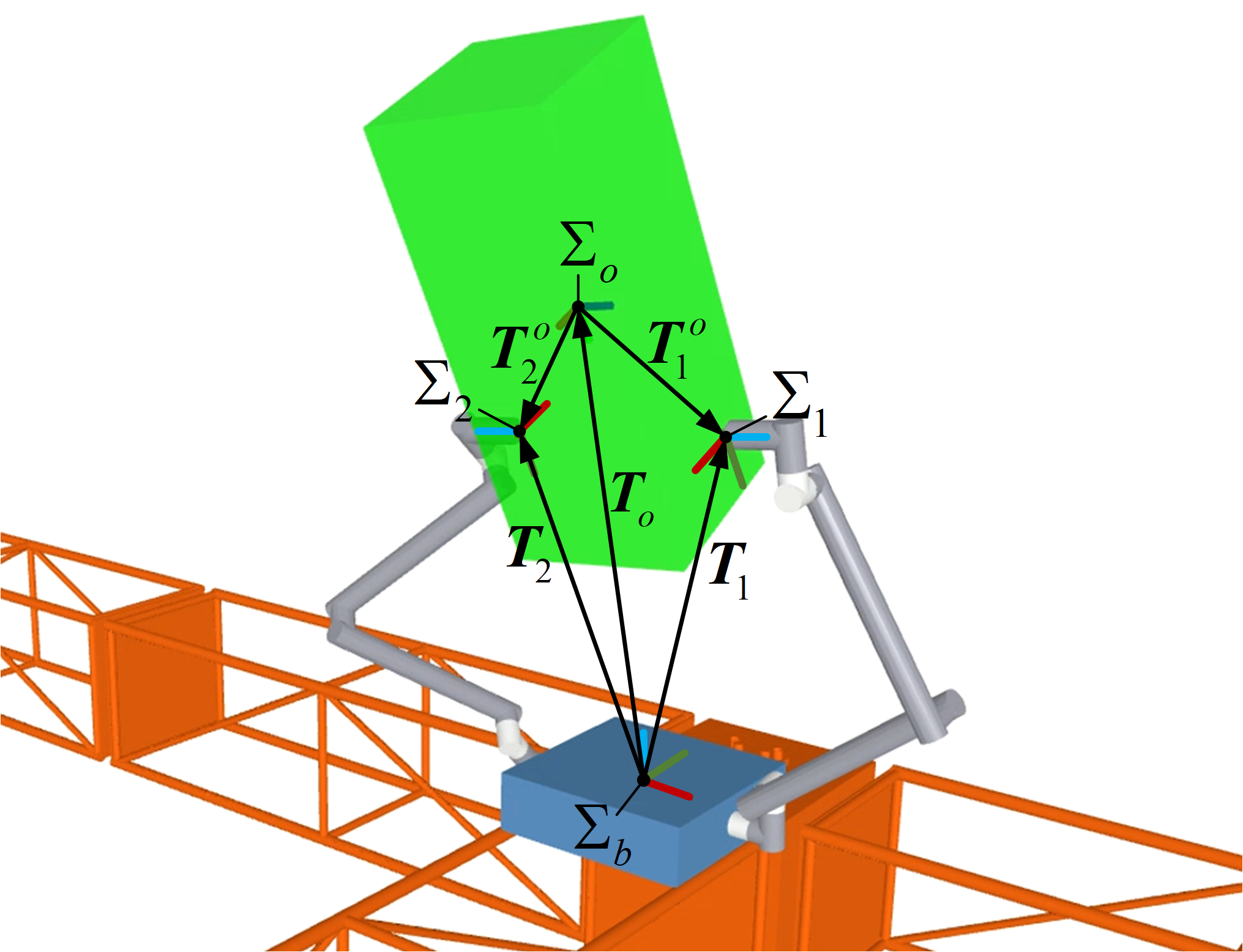} %
	\caption{Diagram of the key coordinate systems and vectors of the dual-arm space manipulator .}
	\label{fig2}
\end{figure}

The coordinate system fixed to the center of mass of the controlled object is represented as ${\sum _o}$. Use ${\boldsymbol{p}_o} \in {\mathbb{R}^3}$, ${\boldsymbol{R}_o} \in SO(3)$ and ${\boldsymbol{T}_o} \in SE(3)$ to denote the position vector, rotation matrix and homogeneous transformation matrix of the ${\sum _o}$. Use $\boldsymbol{r}_i^o \in {\mathbb{R}^3}$, ${\boldsymbol{R}_i^o} \in SO(3)$ and ${\boldsymbol{T}_i^o} \in SE(3)$ to represent the position vector, rotation matrix and homogeneous transformation matrix of the ${\sum _i}$ relative to the ${\sum _o}$. The key coordinate systems and vectors of the dual-arm space manipulator are shown in Fig. \ref{fig2}. In this paper, the superscript on the right of a variable indicates in which coordinate system it is represented. If it is represented in the base coordinate system ${\sum _b}$, the superscript is omitted by default. For the tightly coupled dual-arm space manipulator, we assume that there is no relative motion between the end effectors of each manipulator and the controlled object, that is, ${\boldsymbol{p}_i^o}$, ${\boldsymbol{R}_i^o}$ and ${\boldsymbol{T}_i^o}$ remain unchanged. There is a relationship between the joint angles of the two manipulators:
\begin{equation}
	{\boldsymbol{T}_2}{\left( {{\boldsymbol{q}_2}} \right)^{ - 1}}{\boldsymbol{T}_1}\left( {{\boldsymbol{q}_1}} \right)\boldsymbol{T}_2^1 = {\boldsymbol{I}_{4 \times 4}}
\end{equation}

Use ${\boldsymbol{T}_e}(\boldsymbol{q}) = {\boldsymbol{T}_2}{\left( {{\boldsymbol{q}_2}} \right)^{ - 1}}{\boldsymbol{T}_1}\left( {{\boldsymbol{q}_1}} \right)\boldsymbol{T}_2^1$ represent the constraint deviation of the closed-chain constraint. ${\boldsymbol{T}_e}$ can be converted into position deviation ${\boldsymbol{p}_e}(\boldsymbol{q}) \in {\mathbb{R}^3}$ and attitude deviation ${\boldsymbol{u}_e}(\boldsymbol{q}) \in {\mathbb{R}^3}$ in the form of PRY Euler angles. Let $\boldsymbol{h}(\boldsymbol{q}) = {[\boldsymbol{p}_e^{\rm{T}},\boldsymbol{u}_e^{\rm{T}}]^{\rm{T}}} = \Phi ({\boldsymbol{T}_e})$ represent the deviation vector corresponding to ${\boldsymbol{T}_e}$, and $\Phi( \cdot )$ represent the conversion function between the homogeneous transformation matrix and the deviation vector:
\begin{equation}
	\boldsymbol{h}(\boldsymbol{q}) = {\Phi }({\boldsymbol{T}_e}) = \left[ {\begin{array}{*{20}{c}}
			{{{T}_{1,4}}}\\
			{{{T}_{2,4}}}\\
			{{{T}_{3,4}}}\\
			{{\rm{atan2}}\left( {{{T}_{3,2}},{{T}_{3,3}}} \right)}\\
			{ - {\rm{atan2}}\left( {{{T}_{3,1}}}, \sqrt{T_{3,2}^2 + T_{3,3}^2} \right)}\\
			{{\rm{atan2}}\left( {{{T}_{2,1}},{{T}_{1,1}}} \right)}
	\end{array}} \right]
\end{equation}

Where ${{T}_{a,b}}$ represents the element at the a-th row and b-th column of the homogeneous transformation matrix ${\boldsymbol{T}_e}$. $\boldsymbol{h}(\boldsymbol{q})$ represents the closed-chain constraint of the dual-arm space manipulator. The vector space corresponding to the joint angle vector ${\boldsymbol{q}}$ is called the joint space ${\cal Q}$, comprising obstacle ${{\cal Q}_{{\rm{obs}}}}$ and obstacle-free ${{\cal Q}_{{\rm{free}}}} = {\cal Q}\backslash {{\cal Q}_{{\rm{obs}}}}$. All configurations satisfying the closed-chain constraint form the manifold: ${\cal M} = \left\{ {\boldsymbol{q} \in {\cal Q}|\boldsymbol{h}(\boldsymbol{q}) = {\bf{0}}} \right\}$. The manifold also comprising obstacle ${{\cal M}_{{\rm{obs}}}} = {\cal M} \cap {{\cal Q}_{{\rm{obs}}}}$ and obstacle-free ${{\cal M}_{{\rm{free}}}} = {\cal M} \cap {{\cal Q}_{{\rm{free}}}}$. The motion planning problem is to plan a collision-free motion path $\sigma$ that satisfies the closed-chain constraint, given starting configuration ${\boldsymbol{q}_{{\rm{init}}}}$ and goal configuration ${\boldsymbol{q}_{{\rm{goal}}}}$, such that $\sigma :[0,1] \to {{\cal M}_{{\rm{free}}}}$,$\sigma (0) = {\boldsymbol{q}_{{\rm{init}}}}$, $\sigma (1) = {\boldsymbol{q}_{{\rm{goal}}}}$.

The task space refers to the vector space used to represent the tasks of the manipulator. For the transportation task of the tightly coupled dual-arm space manipulator, two space manipulators will grasp a controlled object simultaneously, and the vector space corresponding to the pose of the controlled object is taken as the task space.

It should be noted that this work focuses on the kinematic motion planning layer of the on-orbit transportation task. The objective is to generate a collision-free reference path that satisfies the closed-chain kinematic constraints between the two manipulators and the manipulated object. The effects of free-floating base dynamics, structural flexibility, sensing uncertainty, and control uncertainty are not explicitly modeled in the present formulation. This assumption is reasonable for scenarios where the dual-arm manipulators are mounted on a large spacecraft whose mass and inertia are much larger than those of the manipulators and the transported module, or where base attitude disturbances are compensated by the spacecraft attitude control system.

\section{Method}

\begin{figure*}[t]
	\centering
	\setlength{\abovecaptionskip}{0.cm}
	\includegraphics[width=0.95\textwidth]{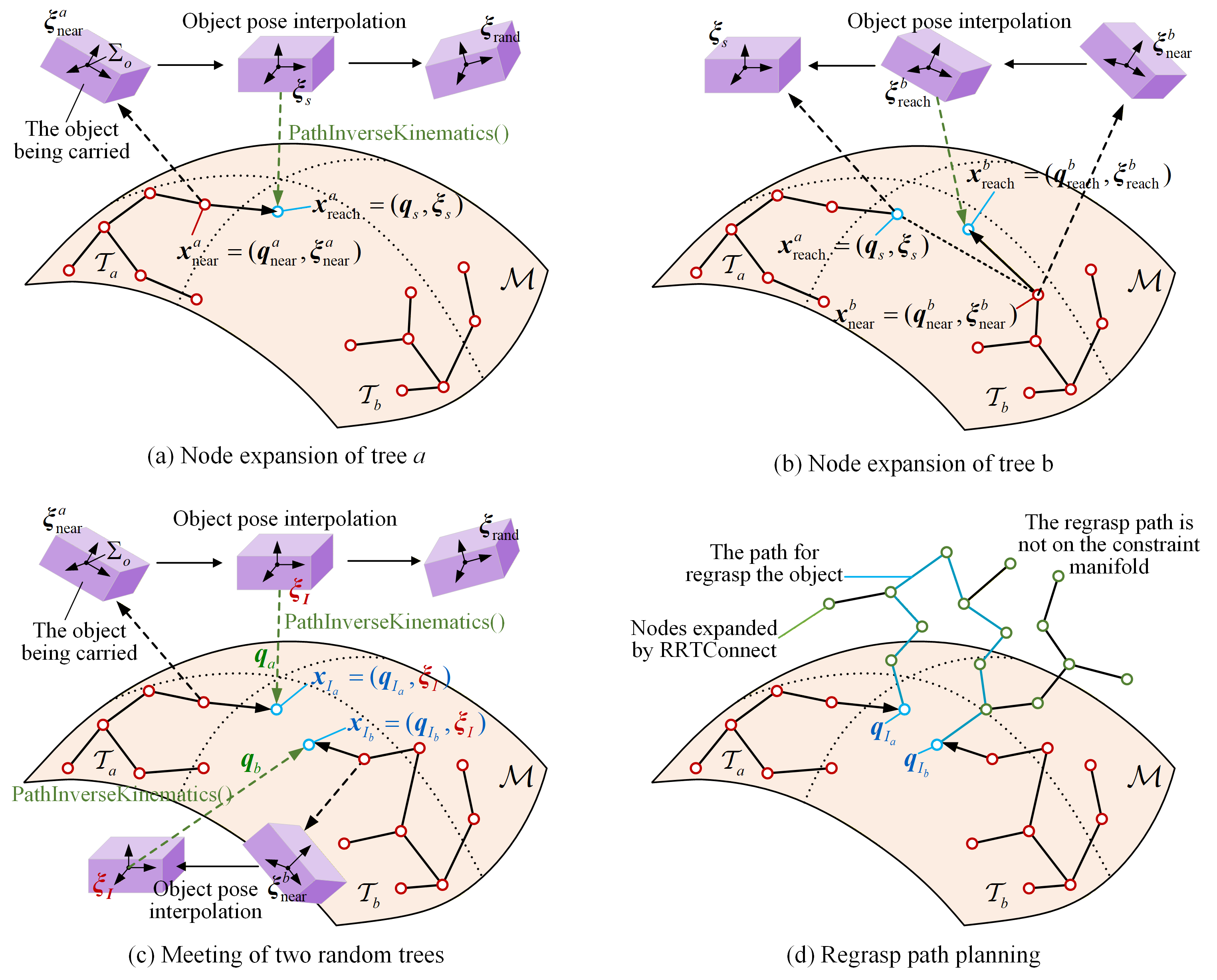} %
	\caption{The schematic diagram of the node expansion process of the TCBiRRT algorithm.}
	\label{fig3}
\end{figure*}

\subsection{The Overall Framework of The Algorithm}

In order to rapidly plan the tightly coupled motion path of the dual-arm space manipulator, the TCBiRRT algorithm is proposed, in which the task space is adopted as the planning space. The pseudo-code of the TCBiRRT algorithm is presented in Algorithm \ref{algorithm1}. The algorithm consists of several key components, including RandomSampleT(), NearestNeighborT(), and ConstrainedExtendT(). RandomSampleT() generates random nodes in the task space. NearestNeighborT() finds the nearest node in the random tree by evaluating the distance metric between nodes. ConstrainedExtendT() expands the random tree on the constrained manifold and generates new nodes.

The results of ConstrainedExtendT() include three situations: \enquote{Trapped}, \enquote{Reached}, and \enquote{Advanced}. If the tree successfully connects to ${\boldsymbol{x}_{{\rm{rand}}}}$, the state \enquote{Reached} is returned. If the random tree cannot connect to ${\boldsymbol{x}_{{\rm{rand}}}}$ but a single-step expansion to ${\boldsymbol{x}_{{\rm{rand}}}}$ is successfully completed, \enquote{Advanced} will be returned. Otherwise, the state \enquote{Trapped} is returned.

During the planning process, a random node ${\boldsymbol{x}_{{\rm{rand}}}}$ is sampled in the planning space firstly. Then, the random tree is expanded towards the ${\boldsymbol{x}_{{\rm{rand}}}}$ on the constraint manifold to obtain the node ${\boldsymbol{x}_{{\rm{reach}}}}$. If the expansion result is not \enquote{Trapped}, expand from the ${\boldsymbol{x}_{{\rm{reach}}}}$ to another random tree until the expansion result is \enquote{Trapped} or connected to another random tree. The expansion process of the random trees is shown in Fig. \ref{fig3}.

If the two random trees do not meet, the Swap() method is used to swap the two trees and repeat the planning process. If the two random trees meet, then plan the regrasp path. If the regrasp path is planned successfully, the ExtractPath() method is used to extract the complete path.

\begin{algorithm}
	\caption{TCBiRRT}
	\label{algorithm1}
	\LinesNotNumbered
	\textbf{Input:} start node ${\boldsymbol{x}_{{\rm{init}}}}$, goal node ${\boldsymbol{x}_{{\rm{goal}}}}$
	
	\textbf{Output:} collision-free motion path satisfying closed-chain constraints
	\LinesNumbered
	
	\nl ${{\cal T}_a}.{\rm{AddRoot}}({\boldsymbol{x}_{{\rm{init}}}})$, ${{\cal T}_b}.{\rm{AddRoot}}({\boldsymbol{x}_{{\rm{goal}}}})$
	
	\nl \While{not timeout}{
		\nl ${\boldsymbol{x}_{{\rm{rand}}}} \leftarrow $ RandomSampleT()
		
		\nl $\boldsymbol{x}_{{\rm{near}}}^a \leftarrow $ NearestNeighborT(${{\cal T}_a}{\rm{, }}{\boldsymbol{x}_{{\rm{rand}}}}$)
		
		\nl $\boldsymbol{x}_{{\rm{reach}}}^a, {\rm{res}}, \boldsymbol{Q}_{\rm{regrasp}} \leftarrow $ ConstrainedExtendT(${{\cal T}_a}{\rm{, }}\boldsymbol{x}_{{\rm{near}}}^a{\rm{, }}{\boldsymbol{x}_{{\rm{rand}}}}$)
		
		\nl \If{\rm{res}$\neq$\rm{Trapped} }{
			\nl $\boldsymbol{x}_{{\rm{near}}}^b \leftarrow $ NearestNeighborT(${{\cal T}_b}{\rm{, }}\boldsymbol{x}_{{\rm{reach}}}^a$)
			
			\nl $\boldsymbol{x}_{{\rm{reach}}}^b, {\rm{res}}, \boldsymbol{Q}_{\rm{regrasp}} \leftarrow $ ConstrainedExtendT(${{\cal T}_b}{\rm{, }}\boldsymbol{x}_{{\rm{near}}}^b{\rm{, }}\boldsymbol{x}_{{\rm{reach}}}^a$)
			
			\nl \While{\rm{res}=\rm{Advanced }}{
				\nl $\boldsymbol{x}_{{\rm{near}}}^b \leftarrow $ NearestNeighborT(${{\cal T}_b}{\rm{, }}\boldsymbol{x}_{{\rm{reach}}}^a$)
				
				\nl $\boldsymbol{x}_{{\rm{reach}}}^b,{\rm{res}}, \boldsymbol{Q}_{\rm{regrasp}} \leftarrow $ ConstrainedExtendT(${{\cal T}_b}{\rm{, }}\boldsymbol{x}_{{\rm{near}}}^b{\rm{, }}\boldsymbol{x}_{{\rm{reach}}}^a$)
				
			}
			
			\nl \If{\rm{res}=\rm{Reached}}{
				
				\nl $\boldsymbol{X} \leftarrow $ ExtractPath(${{\cal T}_a}{\rm{, }}\boldsymbol{x}_{{\rm{reach}}}^a,{{\cal T}_b}{\rm{, }}\boldsymbol{x}_{{\rm{reach}}}^b, \boldsymbol{Q}_{\rm{regrasp}}$)
				
				\nl \textbf{return} $\boldsymbol{X}$
			}
		}
		
		\nl Swap$\left( {{\cal T}_a},{{\cal T}_b} \right) $
		
	}
	\nl \textbf{return} false
	
\end{algorithm}

\begin{algorithm}
	\caption{ConstrainedExtendT}
	\label{algorithm2}
	\textbf{Input:} the near node ${\boldsymbol{x}_{{\rm{near}}}} = ({\boldsymbol{q}_{{\rm{near}}}},{\boldsymbol{\xi }_{{\rm{near}}}})$; the random node ${\boldsymbol{x}_{{\rm{rand}}}} = ({\boldsymbol{q}_{{\rm{rand}}}},{\boldsymbol{\xi }_{{\rm{rand}}}})$; the random tree ${\cal T}$
	
	\textbf{Output:} extended node ${\boldsymbol{x}_s} = ({\boldsymbol{q}_s},{\boldsymbol{\xi }_s})$; results of the extension: res; regrasp path $ \boldsymbol{Q}_{\rm{regrasp}}$
	
	\nl ${\boldsymbol{\xi }_s} \leftarrow {\boldsymbol{\xi }_{{\rm{near}}}}$, ${\boldsymbol{q}_s} \leftarrow {\boldsymbol{q}_{{\rm{near}}}}$
	
	\nl ${\rm{res}} \leftarrow \textit{Trapped}$, $ \boldsymbol{Q}_{\rm{regrasp}} \leftarrow \emptyset$
	
	\nl ${\boldsymbol{\xi }_s} \leftarrow {\boldsymbol{\xi }_{{\rm{near}}}} + \min ({s_\xi },{\left\| {{\boldsymbol{\xi }_{{\rm{rand}}}} - {\boldsymbol{\xi }_{{\rm{near}}}}} \right\|})\frac{{{\boldsymbol{\xi }_{{\rm{rand}}}} - {\boldsymbol{\xi }_{{\rm{near}}}}}}{{{{\left\| {{\boldsymbol{\xi }_{{\rm{rand}}}} - {\boldsymbol{\xi }_{{\rm{near}}}}} \right\|}}}}$
	
	\nl ${{\boldsymbol{P}_o},{\boldsymbol{O}_o}} \leftarrow$ Interpolate(${\boldsymbol{\xi }_{{\rm{near}}}},{\boldsymbol{\xi }_s}$)
	
	\nl ${{\boldsymbol{Q}_{{\rm{near - s}}}},{\boldsymbol{q}_s}} \leftarrow$ PathInverseKinematics(${{\boldsymbol{P}_o},{\boldsymbol{O}_o},{\boldsymbol{q}_{{\rm{near}}}}}$)
	
	\nl \If{${\boldsymbol{q}_s} \neq$ \rm{NULL and CollisionFree}(${\boldsymbol{Q}_{{\rm{near - s}}}}$)}{
		
		\nl \eIf{${\boldsymbol{\xi }_s}\boldsymbol{ = }{\boldsymbol{\xi }_{{\rm{rand}}}}$ \rm{and} ${\boldsymbol{q}_{{\rm{rand}}}}\neq$ NULL}{
			
			\nl $Success, \boldsymbol{Q}_{\rm{regrasp}} \leftarrow $ RRTConnect(${{\boldsymbol{q}_s}, \boldsymbol{q}_{{\rm{rand}}}}$)
			
			\nl \eIf{Success}{
				\nl ${\boldsymbol{x}_s} \leftarrow ({\boldsymbol{q}_s},{\boldsymbol{\xi }_s})$
				
				\nl ${\cal T}.{\rm{AddNode}}({\boldsymbol{x}_s})$
				
				\nl ${\cal T}.{\rm{AddEdge}}({\boldsymbol{x}_{{\rm{near}}}},{\boldsymbol{x}_s})$
				
				\nl \textbf{return} ${\boldsymbol{x}_s, \rm{res} \leftarrow Reached, \boldsymbol{Q}_{\rm{regrasp}}}$
				
			}
			{
				
				\nl \textbf{return} ${\boldsymbol{x}_s, \rm{res} \leftarrow Trapped, \boldsymbol{Q}_{\rm{regrasp}}}$
				
			}
			
		}{
			
			\nl ${\boldsymbol{x}_s} \leftarrow ({\boldsymbol{q}_s},{\boldsymbol{\xi }_s})$
			
			\nl ${\cal T}.{\rm{AddNode}}({\boldsymbol{x}_s})$
			
			\nl ${\cal T}.{\rm{AddEdge}}({\boldsymbol{x}_{{\rm{near}}}},{\boldsymbol{x}_s})$

		}
		
		\nl \eIf{${\boldsymbol{\xi }_s}\boldsymbol{ = }{\boldsymbol{\xi }_{{\rm{rand}}}}$}{

			\nl \textbf{return} ${\boldsymbol{x}_s, \rm{res} \leftarrow Reached, \boldsymbol{Q}_{\rm{regrasp}}}$
			
		}
		{
			
			\nl \textbf{return} ${\boldsymbol{x}_s, \rm{res} \leftarrow Advanced, \boldsymbol{Q}_{\rm{regrasp}}}$
			
		}
		
	}
	
	\nl \textbf{return} ${\boldsymbol{x}_s \leftarrow ({\boldsymbol{q}_s},{\boldsymbol{\xi }_s}), \rm{res} \leftarrow Trapped, \boldsymbol{Q}_{\rm{regrasp}}}$
	
\end{algorithm}

\begin{algorithm}
	\caption{PathInverseKinematics}
	\label{algorithm3}
	\textbf{Input:} the position sequence of object ${\boldsymbol{P}_o} = \{ {}^j{\boldsymbol{p}_o}\} _{j = 0}^N$; the orientation sequence of object  ${\boldsymbol{O}_o} = \{ {}^j{\boldsymbol{R}_o}\} _{j = 0}^N$; initial joint angle  ${\boldsymbol{q}_{{\rm{init}}}} = {[\boldsymbol{q}_1^{\rm{T}},\boldsymbol{q}_2^{\rm{T}}]^{\rm{T}}}$
	
	\textbf{Output:} sequence of joint angles $\boldsymbol{Q}{\rm{ = \{ }}{}^j\boldsymbol{q}{\rm{\} }}_{j = 0}^N$, and the last reached joint angle $\boldsymbol{q}$

	\nl $\boldsymbol{Q} \leftarrow \emptyset ;{}^j\boldsymbol{q} \leftarrow {\boldsymbol{q}_{{\rm{init}}}}; Success \leftarrow$ True

	\nl \For{j = \rm{0} : $N$}{
		
		\nl \For{$i=1:2$}{
			
			\nl ${}^j{\boldsymbol{p}_{g_i}}
			\leftarrow
			{}^j{\boldsymbol{p}_o}
			+
			{}^j{\boldsymbol{R}_o}\boldsymbol{r}_i^o$
			
			\nl ${}^j{\boldsymbol{R}_{g_i}}
			\leftarrow
			{}^j{\boldsymbol{R}_o}\boldsymbol{R}_i^o$
		}
		
		\nl ${}^j{\boldsymbol{p}_g} \leftarrow \{ {{}^j{\boldsymbol{p}_{g_i}}}\} _{i = 1}^2$
		
		\nl ${}^j{\boldsymbol{R}_g} \leftarrow \{ {{}^j{\boldsymbol{R}_{g_i}}}\} _{i = 1}^2$

		\nl ${}^j{\boldsymbol{q}_g} \leftarrow IK({}^j\boldsymbol{q},{}^j{\boldsymbol{p}_g},{}^j{\boldsymbol{R}_g})$
		
		\nl \If{${}^j{\boldsymbol{q}_g}$ \rm{is NULL}}{
			
			\nl $Success \leftarrow $False
			
			\nl \textbf{break}
			
		}
		
		\nl $\boldsymbol{Q}$.append(${}^j{\boldsymbol{q}_g}$)
		
		\nl ${}^j\boldsymbol{q} \leftarrow {}^j{\boldsymbol{q}_g}$
		
	}
	
	\nl \If{Success }{
		
		\nl \textbf{return $\boldsymbol{Q}, {}^N{\boldsymbol{q}_g}$}
		
	}
	
	\nl \textbf{return $\boldsymbol{Q}$}, NULL

\end{algorithm}

\begin{algorithm}
	\caption{InverseKinematics(IK)}
	\label{algorithm4}
	\textbf{Input:} initial joint angle for all manipulators $\boldsymbol{q} = {[\boldsymbol{q}_1^{\rm{T}},\boldsymbol{q}_2^{\rm{T}}]^{\rm{T}}}$, desired position of the end-effectors ${\boldsymbol{p}_g} = \{ {\boldsymbol{p}_{{g_i}}}\} _{i = 1}^2$, desired rotation matrix of the end-effectors ${\boldsymbol{R}_g} = \{ {\boldsymbol{R}_{{g_i}}}\} _{i = 1}^2$  
	
	\textbf{Output:} desired joint angle for all manipulators ${\boldsymbol{q}_g} = {[\boldsymbol{q}_{{g_1}}^{\rm{T}},\boldsymbol{q}_{{g_2}}^{\rm{T}}]^{\rm{T}}}$

	\nl \For{i = \rm{1 : 2}}{
		
		\nl ${\boldsymbol{q}_{{g_i}}} \leftarrow $ NULL
		
		\nl \For{k = \rm{1} : $K_{max}$}{
			
			\nl ${\boldsymbol{p}_{{t_i}}},{\boldsymbol{R}_{{t_i}}} \leftarrow F{K_i}\left( {{\boldsymbol{q}_i}} \right)$
			
			\nl ${\boldsymbol{e}_{{p_i}}} \leftarrow {\boldsymbol{p}_{{g_i}}} - {\boldsymbol{p}_{{t_i}}}$
			
			\nl ${\boldsymbol{e}_{{o_i}}} \leftarrow {\boldsymbol{R}_{{g_i}}}\boldsymbol{\varphi }\left( {{{\left( {{\boldsymbol{R}_{{t_i}}}} \right)}^{\rm{T}}}{\boldsymbol{R}_{{g_i}}}} \right)$
			
			\nl ${\boldsymbol{\dot q}_i} \leftarrow \boldsymbol{J}_{{\boldsymbol{q}_i}}^\dag \boldsymbol{K}{\left[ {\boldsymbol{e}_{{p_i}}^{\rm{T}},\boldsymbol{e}_{{o_i}}^{\rm{T}}} \right]^{\rm{T}}}$
			
			\nl ${\boldsymbol{q}_i} \leftarrow {\boldsymbol{q}_i} + \lambda {\boldsymbol{\dot q}_i}$
			
			\nl \If{$\left\| {{\boldsymbol{e}_{{p_i}}}} \right\| < {\varepsilon _p}$ \rm{and} $\left\| {{\boldsymbol{e}_{{o_i}}}} \right\| < {\varepsilon _o}$}{
				
				\nl ${\boldsymbol{q}_{{g_i}}} \leftarrow {\boldsymbol{q}_i}$
				
			}
			
		}
		
		\nl \If{${\boldsymbol{q}_{{g_i}}}=$ \rm{NULL}}{
			
			\nl \textbf{return} NULL
			
		}
		
	}
	
	\nl ${\boldsymbol{q}_g} \leftarrow {[\boldsymbol{q}_{{g_1}}^{\rm{T}},\boldsymbol{q}_{{g_2}}^{\rm{T}}]^{\rm{T}}}$

	\nl \textbf{return} ${\boldsymbol{q}_g}$

\end{algorithm}

\subsection{Expansion of Nodes in The Task Space}

\begin{figure*}[t]
	\centering
	\setlength{\abovecaptionskip}{0.cm}
	\includegraphics[width=0.95\textwidth]{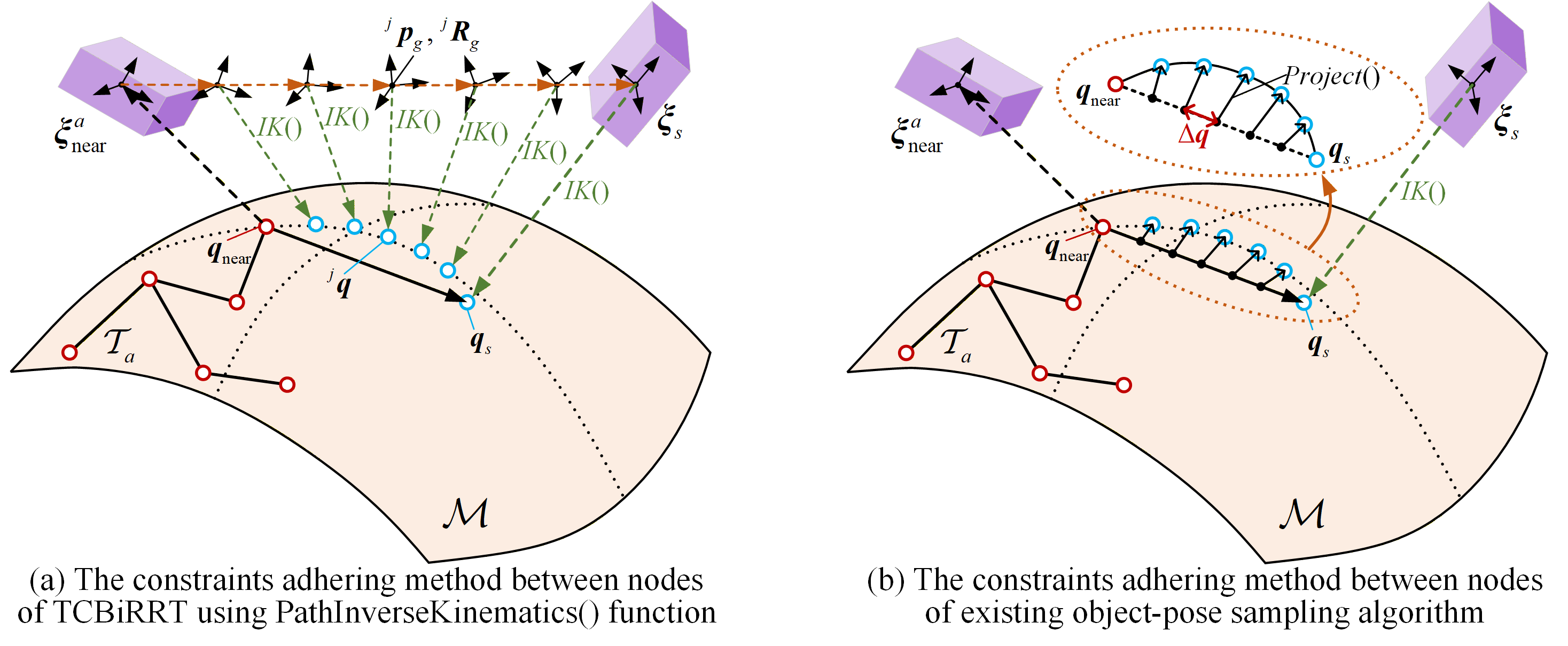} %
	\caption{Comparison with existing planning algorithms based on object pose sampling.}
	\label{fig4}
\end{figure*}

Each node in the random tree of the TCBiRRT algorithm contains two components: the joint angle vector $\boldsymbol{q}$ of the manipulator in the configuration space and the pose $\boldsymbol{\xi}= {[\boldsymbol{p}^{\rm{T}},\boldsymbol{u}^{\rm{T}}]^{\rm{T}}} \in {\mathbb{R}^6}$ of the manipulated object in the task space. In the $\boldsymbol{\xi}$, $\boldsymbol{p}$ and $\boldsymbol{u}$ represent the position and orientation of the object, respectively. The orientation $\boldsymbol{u}$ is expressed in terms of Euler angles. Accordingly, the initial node and goal node are defined as ${\boldsymbol{x}_{{\rm{init}}}} = ({\boldsymbol{q}_{{\rm{init}}}},{\boldsymbol{\xi }_{{\rm{init}}}})$ and ${\boldsymbol{x}_{{\rm{goal}}}} = ({\boldsymbol{q}_{{\rm{goal}}}},{\boldsymbol{\xi }_{{\rm{goal}}}})$, respectively.

When the task space is used as the planning space, the RandomSampleT() generates random nodes by uniformly sampling the pose of the manipulated object. Specifically, the pose ${\boldsymbol{\xi }_{{\rm{rand}}}}$ is sampled within a hyper-rectangular region in the 6-dimensional task space. Let ${\boldsymbol{\xi }_l}$ and ${\boldsymbol{\xi }_h}$ denote the lower and upper bounds of the sampling space, respectively. The sampling space is defined as: ${\rm{\{ }}\boldsymbol{\xi } = {\rm{(}}{\xi_1}{\rm{,}}{\xi_2}{\rm{,}} \cdots {\rm{,}}{\xi _6}{\rm{)|}}{\xi _{li}} \le {\xi _i} \le {\xi _{hi}}{\rm{,}}i{\rm{ = 1,2,}} \cdots {\rm{,6\} }}$.

The node corresponding to ${\boldsymbol{\xi }_{{\rm{rand}}}}$ is denoted as ${\boldsymbol{x}_{{\rm{rand}}}} = ({\boldsymbol{q}_{{\rm{rand}}}},{\boldsymbol{\xi }_{{\rm{rand}}}})$. Since only the object pose is sampled, ${\boldsymbol{q}_{{\rm{rand}}}}$ is undefined at this stage. The NearestNeighborT() finds the nearest node ${\boldsymbol{x}_{{\rm{near}}}} = ({\boldsymbol{q}_{{\rm{near}}}},{\boldsymbol{\xi }_{{\rm{near}}}})$ in the random tree by minimizing the Euclidean distance $\boldsymbol{d} = {\left\| {{\boldsymbol{\xi }_{{\rm{rand}}}} - {\boldsymbol{\xi }_{{\rm{near}}}}} \right\|_2}$ in the task space. The ConstrainedExtendT() is then applied to extend the tree from ${\boldsymbol{x}_{{\rm{near}}}}$ toward ${\boldsymbol{x}_{{\rm{rand}}}}$ and generate new nodes. The pseudo-code of this process is provided in Algorithm \ref{algorithm2}.

The new object pose ${\boldsymbol{\xi }_s}$ is then computed by extending from ${\boldsymbol{\xi }_{{\rm{near}}}}$ toward ${\boldsymbol{\xi }_{{\rm{rand}}}}$ with step size ${s_\xi}$:
\begin{equation}
	{\boldsymbol{\xi }_s} \leftarrow {\boldsymbol{\xi }_{{\rm{near}}}} + \min ({s_\xi},{\left\| {{\boldsymbol{\xi }_{{\rm{rand}}}} - {\boldsymbol{\xi }_{{\rm{near}}}}} \right\|})\frac{{{\boldsymbol{\xi }_{{\rm{rand}}}} - {\boldsymbol{\xi }_{{\rm{near}}}}}}{{{{\left\| {{\boldsymbol{\xi }_{{\rm{rand}}}} - {\boldsymbol{\xi }_{{\rm{near}}}}} \right\|}}}}
	\label{eq:1}
\end{equation}

Interpolation is then performed in the task space. In this paper, the object pose is interpolated using rotational exponential coordinates $\boldsymbol{\varphi} \in {\mathbb{R}^3}$. Since the orientations are represented by Euler angles in ${\boldsymbol{\xi }_{{\rm{near}}}}$ and ${\boldsymbol{\xi }_s}$, they are first converted into rotation matrices ${\boldsymbol{R}_{{\rm{near}}}}$ and ${\boldsymbol{R}_s}$, and then converted into exponential coordinates ${\boldsymbol{\varphi }_{{\rm{near}}}}$ and ${\boldsymbol{\varphi }_s}$. The conversion from rotation matrices to exponential coordinates is presented in the Appendix. The interpolation of the object pose is defined as:
\begin{equation}
	\left\{ \begin{array}{l}
		{}^j{\boldsymbol{p}_o} = {\boldsymbol{p}_{{\rm{near}}}} + j\left( {{\boldsymbol{p}_s} - {\boldsymbol{p}_{{\rm{near}}}}} \right){\rm{/}}N\\
		{}^j{\boldsymbol{\varphi }_o} = {\boldsymbol{\varphi }_{{\rm{near}}}} + j\left( {{\boldsymbol{\varphi }_s} - {\boldsymbol{\varphi }_{{\rm{near}}}}} \right){\rm{/}}N
	\end{array} \right.j = 1,2, \cdots ,N
\end{equation}

Here, $N$ denotes the number of interpolation points. The interpolated orientation ${}^j{\boldsymbol{\varphi }_o}$ is converted into the rotation matrix ${}^j{\boldsymbol{R}_o}$. Consequently, the position sequence ${\boldsymbol{P}_o} = \{ {}^j{\boldsymbol{p}_o}\} _{j = 0}^N$ and orientation sequence ${\boldsymbol{O}_o} = \{ {}^j{\boldsymbol{R}_o}\} _{j = 0}^N$ are obtained. To verify feasibility, the corresponding continuous joint angle sequence ${\boldsymbol{Q}_{{\rm{near - s}}}} = {\rm{\{ }}{}^j\boldsymbol{q}{\rm{\} }}_{j = 0}^N$ must be computed. For this purpose, a path inverse kinematics algorithm PathInverseKinematics() is proposed. 

The pseudo-code of PathInverseKinematics() is shown in Algorithm \ref{algorithm3}. At each waypoint $j$, the object pose ${}^j{\boldsymbol{p}_o}$ and ${}^j{\boldsymbol{R}_o}$ are used to compute the desired end-effector poses ${}^j{\boldsymbol{p}_g} = \{ {}^j{\boldsymbol{p}_{{g_i}}}\} _{i = 1}^2$ and ${}^j{\boldsymbol{R}_g} = \{ {}^j{\boldsymbol{R}_{{g_i}}}\} _{i = 1}^2$. The inverse kinematics solver $IK()$ is then applied to obtain joint angle ${}^j{\boldsymbol{q}_g}$ at waypoint $j$. The pseudo-code of $IK()$ is shown in Algorithm \ref{algorithm4}.

We use the numerical method to implement the inverse kinematics solver $IK()$. The desired pose of the end-effector of the manipulator $i$ is represented by ${\boldsymbol{T}_{{g_i}}}$, and the initial joint angle is represented by ${\boldsymbol{q}_i}$. The corresponding position and rotation matrix of ${\boldsymbol{T}_{{g_i}}}$ are denoted as ${\boldsymbol{p}_{{g_i}}}$ and ${\boldsymbol{R}_{{g_i}}}$ respectively. The initial position and initial rotation matrix of the end-effector are respectively denoted as ${\boldsymbol{p}_{{t_i}}}$ and ${\boldsymbol{R}_{{t_i}}}$. The position error and attitude error are represented as ${\boldsymbol{e}_{{p_i}}}$ and ${\boldsymbol{e}_{{o_i}}}$ respectively. The ${\boldsymbol{e}_{{o_i}}}$ is calculated based on the rotational exponential coordinates.

\begin{equation}
	{\boldsymbol{e}_{{p_i}}} = {\boldsymbol{p}_{{g_i}}} - {\boldsymbol{p}_{{t_i}}}
\end{equation}
\begin{equation}
	{\boldsymbol{e}_{{o_i}}} = {\boldsymbol{R}_{{g_i}}}\boldsymbol{\varphi }\left( {{{\left( {{\boldsymbol{R}_{{t_i}}}} \right)}^{\rm{T}}}{\boldsymbol{R}_{{g_i}}}} \right)
\end{equation}

In the above formulation, $\boldsymbol{\varphi }\left( \cdot \right)$ denotes the mapping from a rotation matrix to its corresponding exponential coordinates, as detailed in the Appendix. The joint velocity ${\boldsymbol{\dot q}_i}$ is then computed using the pseudo-inverse of the Jacobian matrix of manipulator $i$:
\begin{equation}
	{\boldsymbol{\dot q}_i}\boldsymbol{ = J}_{{\boldsymbol{q}_i}}^\dag \boldsymbol{K}{\left[ {\boldsymbol{e}_{{p_i}}^{\rm{T}},\boldsymbol{e}_{{o_i}}^{\rm{T}}} \right]^{\rm{T}}}
\end{equation}

Here, $\boldsymbol{K}$ represents the gain matrix. The joint angle of manipulator $i$ is then updated as:
\begin{equation}
	{\boldsymbol{q}_i} = {\boldsymbol{q}_i} + \lambda {\boldsymbol{\dot q}_i}
\end{equation}

Where $\lambda \in {\mathbb{R}^{\rm{ + }}}$ is the step size. By iteratively applying the process from (3) to (6), the end-effector pose gradually converges to the desired target pose. If within a predefined maximum number of iterations $K_{max}$, the position error $\lVert{\boldsymbol{e}_{{p_i}}} \rVert$ is less than the threshold ${\varepsilon _p}$ and the orientation error $\lVert {\boldsymbol{e}_{{o_i}}} \rVert$ is less than the threshold ${\varepsilon _o}$, the solution is considered valid, and the joint angle vector ${\boldsymbol{q}_{{g_i}}}$ corresponding to ${\boldsymbol{T}_{{g_i}}}$ is successfully obtained.

In order to maintain the continuity of the joint motion, the joint angle ${}^j{\boldsymbol{q}_g}$ corresponding to the waypoint ${}^j{\boldsymbol{p}_o}$ and ${}^j{\boldsymbol{R}_o}$ is used as the initial joint angle to calculate the next waypoint ${}^{j + 1}{\boldsymbol{q}_g}$. If the complete joint sequence ${\boldsymbol{Q}_{{\rm{near-s}}}}$ is successfully obtained, collision checking is performed sequentially on all configurations. The node expansion is regarded as successful only if all configurations in ${\boldsymbol{Q}_{{\rm{near-s}}}}$ are collision-free.

It should be noted that object-pose sampling has been used in constrained PRM algorithm \cite{c15}. In such algorithm, object-pose sampling is used to generate roadmap nodes, whereas the feasibility of the continuous constrained motion between two object poses is not directly guaranteed during sampling. In such algorithm, the constraints adhering between nodes is achieved through joint-space interpolation and projection. In contrast, each edge generated by TCBiRRT corresponds to a continuous object-pose interpolation path, and the associated joint-space trajectory is recovered and verified by the proposed PathInverseKinematics procedure. Therefore, the proposed planner explores the low-dimensional object task space while maintaining edge-level feasibility on the closed-chain constraint manifold. Compared with the existing object-pose sampling algorithm, this approach can increase the success rate of node expansion and thus find feasible path more quickly. The comparison between the TCBiRRT algorithm and existing object-pose sampling algorithm is shown in Fig.~\ref{fig4}.

\subsection{Regrasp the Object}

The solution of the PathInverseKinematics() depends not only on the desired end-effector pose but also on the initial joint angle. The initial joint angle is also called the seed vector. Different seed vectors may lead to different solutions.

When the two random trees in the TCBiRRT algorithm meet, the same object pose ${\boldsymbol{\xi}_I}$ will correspond to two different joint angles: ${\boldsymbol{q}_a},{\boldsymbol{q}_b}$. Specifically, $\boldsymbol{q}_a$ is obtained using the joint angle of the parent node of ${{\boldsymbol{x}_I}_a}$ in tree ${{\cal T}_a}$ as the seed vector, while $\boldsymbol{q}_b$ is obtained using the corresponding parent node of ${{\boldsymbol{x}_I}_b}$ in tree ${{\cal T}_b}$, as illustrated in Fig. \ref{fig3} (c).

Therefore, it is necessary to plan a motion that connects $\boldsymbol{q}_a$ and $\boldsymbol{q}_b$. The motion between $\boldsymbol{q}_a$ and $\boldsymbol{q}_b$ can be interpreted as a regrasp process of the manipulated object. This regrasp process corresponds to an unconstrained motion planning problem, the classical RRTConnect \cite{c8} algorithm is adopted to compute the regrasp path. If a feasible regrasp path is found, the planning process terminates successfully. Otherwise, the algorithm continues expanding the tree to search for alternative connections.

In practical execution, the RRTConnect algorithm sequentially plans the movements of two space manipulators. When one manipulator changes its configuration, the other manipulator maintains the object pose. If both manipulators need to change their configurations, the two regrasp motions are planned and executed one after another. Therefore, both manipulators are not allowed to release the object simultaneously, and the object is assumed to remain stationary during the regrasp process. Although RRTConnect is used as an unconstrained planner during the regrasp process, the term “unconstrained” only means that the closed-chain kinematic constraint is not imposed while connecting the two joint configurations. Collision constraints are still explicitly checked. During regrasp planning, the object pose is fixed, and all intermediate configurations generated by RRTConnect are checked for collision. The collision detection includes collisions between the two manipulators, collisions between the manipulators and obstacles, collisions between the manipulators and the fixed object, and collisions between the object and obstacles. The intended contact between the end-effectors and the grasping regions of the object is excluded from collision checking, whereas unintended contacts between manipulator links and the object are treated as collisions. A regrasp path is accepted only when all intermediate configurations are collision-free.

\subsection{Computational Complexity}

The computational cost of one TCBiRRT extension consists of nearest-neighbor search, task-space interpolation, path inverse kinematics, and collision checking. Suppose that the current random tree contains $V$ nodes and that a naive nearest-neighbor search is used. The nearest-neighbor query has complexity $O(V)$. For each extension, the object path is discretized into $N$ interpolation waypoints. At each waypoint, inverse kinematics is solved for two manipulators. If the maximum number of IK iterations is $K_{\max}$ and the cost of one Jacobian-based update for one manipulator is denoted as $C_{IK}$, the inverse-kinematics cost of one extension is approximately $O(2NK_{\max}C_{IK})$. Collision checking is then performed along the generated joint sequence, leading to a cost of $O(NC_{col})$, where $C_{col}$ denotes the cost of one collision query. Therefore, the overall cost of one extension can be written as:

\begin{equation}
O\left(V + 2NK_{\max}C_{IK} + NC_{col}\right)
\end{equation}

In practice, the dominant cost comes from inverse kinematics and collision checking. The proposed method improves efficiency because it reduces the number of unproductive samples and projection attempts. Instead of sampling in the high-dimensional joint space and projecting configurations onto the constraint manifold, TCBiRRT samples directly in the object task space, where each sample corresponds to a physically meaningful object motion. The use of the previous waypoint solution as the seed of the next inverse-kinematics query further improves convergence and helps generate continuous joint trajectories efficiently.

\section{Simulation Test}

\subsection{Environment Settings}

\begin{figure*}
	\centering
	\includegraphics[width=0.95\textwidth]{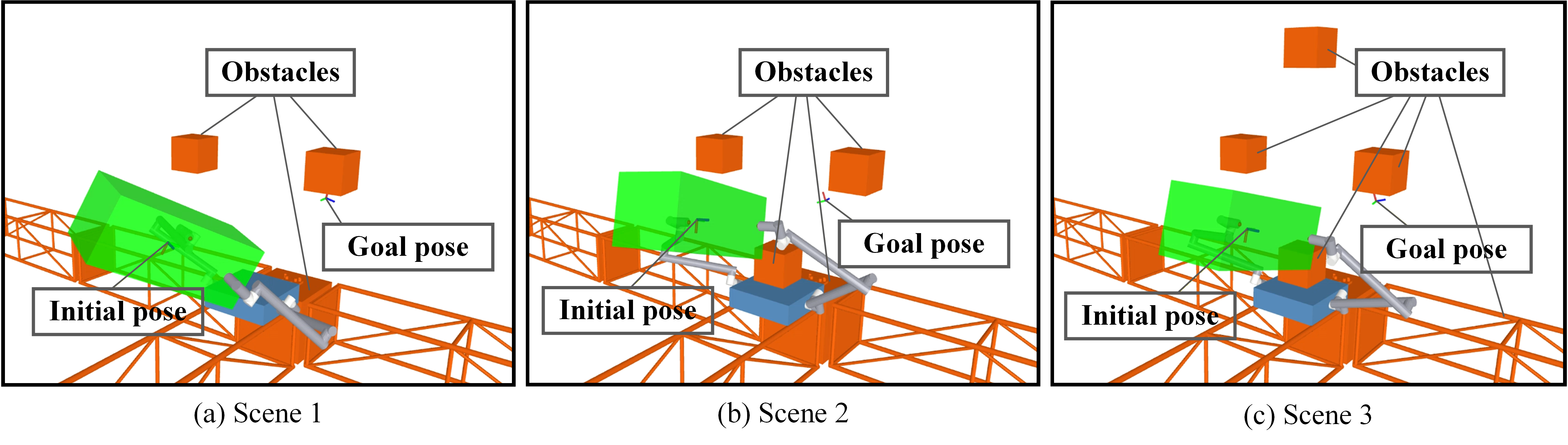}
	\caption{Three partial assembly scenes of the space solar power station in the MuJoCo simulator.}
	\label{fig5}
\end{figure*}

\begin{figure*}
	\centering
	\includegraphics[width=0.95\textwidth]{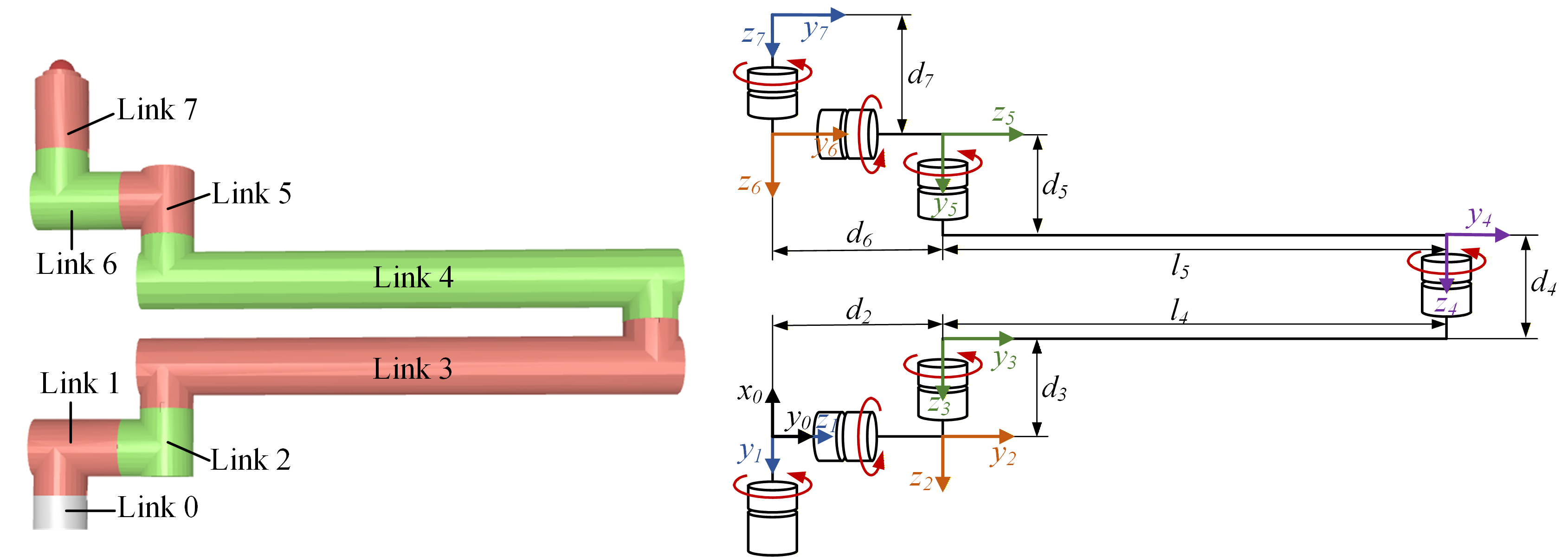}
	\caption{The link coordinate system of the space manipulators.}
	\label{fig6}
\end{figure*}

Three partial assembly scenes of the space solar power station are constructed in the MuJoCo simulator, as illustrated in Fig. \ref{fig5}. The three scenes contain different numbers of obstacles, corresponding to varying levels of motion planning difficulty. In each scene, the red structures and cuboids are obstacles. The green cuboid is the object to be manipulated, with dimensions of 2 m in length, 0.8 m in width, and 0.8 m in height. The link coordinate system of the space manipulators is shown in Fig. \ref{fig6}. The two manipulators share identical D-H parameters, which are listed in Table \ref{table:1}.  The kinematic model of the space manipulators and the collision detection program are implemented using the MuJoCo simulator, and the motion planning algorithms were implemented in Python.

The objective of the planning task is to generate a collision-free motion path for the tightly coupled dual-arm space manipulator given the initial pose ${\boldsymbol{T}_{os}}$ and goal pose ${\boldsymbol{T}_{og}}$ of the manipulated object. To improve the diversity of test cases, a randomized initialization strategy is adopted. Specifically, a nominal initial pose $\boldsymbol{T}_{os}^{init}$ and a nominal goal pose $\boldsymbol{T}_{og}^{init}$ are first defined. Before each planning trial, random pose perturbations are generated and applied to both poses, resulting in ${\boldsymbol{T}_{os}} = \boldsymbol{T}_{os}^{init}{\boldsymbol{T}_{e1}}$, ${\boldsymbol{T}_{og}} = \boldsymbol{T}_{og}^{init}{\boldsymbol{T}_{e2}}$. Each pose perturbation ${\boldsymbol{T}_e}$ consists of a position component and an orientation component. The position perturbation ${\boldsymbol{p}_e} \in {\mathbb{R}^3}$ is sampled uniformly within the range $[-0.2, 0.2]$ m for each dimension. The orientation perturbation is represented by Euler angles ${\boldsymbol{e}_u} \in {\mathbb{R}^3}$, with each component sampled within $[-0.5, 0.5]$ rad. The corresponding initial joint configuration ${\boldsymbol{q}_s}$ and goal joint configuration ${\boldsymbol{q}_g}$ are obtained using the inverse kinematics solver $IK()$. All simulations are conducted on a workstation equipped with an Intel i9-10900K CPU and an Nvidia GeForce RTX 3090 GPU.

\begin{table}
	\centering
	\setlength{\abovecaptionskip}{0.cm}
	\caption{The D-H parameters of space manipulators}
	\label{table:1}
	\begin{tabular}{c c c c c} 
		\toprule 
		Links & ${\theta_i}$/($^{\circ}$) & ${\alpha_i}$/($^{\circ}$) & ${d_i}$/m & ${a_i}$/m \\
		\midrule 
		
		1	& ${\theta_1}$	& 90 & 0 & 0\\
		2 	& ${\theta_2}$	& 90 & 0.24 & 0\\
		3	& ${\theta_3}$	& 0 & -0.17 & 0\\
		4	& ${\theta_4}$	& 0 & -0.18 & 1.08\\
		5	& ${\theta_5}$	& 90 & -0.17 & 1.08\\
		6	& ${\theta_6}$	& 90 & -0.24 & 0\\
		7	& ${\theta_7}$ & 0 & -0.1 & 0\\
		
		\bottomrule 
	\end{tabular}
\end{table}

\subsection{Algorithm Parameters}

The parameters of all the algorithms tested in this paper are as follows:

\begin{enumerate}
	\def\labelenumi{\arabic{enumi})}
	\item
	\textbf{TCBiRRT}\\
	The algorithm proposed in this paper. The node expansion step size ${s_\xi}$ is set to 0.6. Since $\boldsymbol{\xi}$ contains both translational components in meters and rotational components in radians, $s_\xi=0.6$ is a step size parameter in the mixed task-space pose metric. The sampling space boundaries ${\boldsymbol{\xi }_l}$ and ${\boldsymbol{\xi }_h}$ are defined as [-2, -3, 0, -3.14, -3.14, -3.14] and [2, 3, 4, 3.14, 3.14, 3.14], respectively. The number of interpolation points $N$ for task-space discrete collision checking is set to 5, and joint-space regrasp uses 0.2 rad resolution.
	
	\item
	\textbf{CBiRRT2} \cite{c17}\\
	The sampling-based constrained motion planning algorithm that node expansion in joint space, where constraints are enforced using the Newton-Raphson projection. The maximum edge length between nodes is set to 0.6 rad, and the step size for collision checking is set to 0.2 rad.
	
	\item
	\textbf{Precomputed Graph} \cite{c20}\\
	This method extends CBiRRT by constructing a precomputed graph to approximate the constraint manifold. In each scene, 10,000 configurations satisfying the constraints are generated offline to form the graph. During planning, new nodes are sampled directly from this graph.
	
	\item
	\textbf{Latent Sampling} \cite{c22}\\
	This method employs a Conditional Variational Autoencoder (CVAE) \cite{c32} to learn the distribution of configurations on the constraint manifold and maps them into a low-dimensional latent space. During planning, latent vectors are sampled and decoded to efficiently generate configurations near the constraint manifold. For each scene, 10,000 valid configurations are generated for training, and the dataset is augmented using tangent space \cite{c23}. The encoder and decoder of CVAE are both fully connected neural networks with two hidden layers of 512 neurons each, using linear activation functions. The batch size and learning rate are set to 512 and 0.001, respectively, and the latent space dimension is 8. The training process runs for 1000 epochs and requires approximately 8 minutes.
	
	\item
	\textbf{Constrained PRM} \cite{c15}\\
	This probabilistic roadmap method handles closed-chain constraints by sampling the pose of the manipulated object. The sampling range is identical to that used in TCBiRRT. The collision checking step size is set to 0.2 rad.
	
	\item
	\textbf{LCBiRRT} \cite{c23}\\
	This method performs planning in a learned latent space. The latent space is obtained through the CVAE model, which is the same as the Latent Sampling method. The obstacles are represented by voxels and encoded using neural networks. The voxel encoding neural network consists of two hidden layers with 512 neurons each and uses Leaky ReLU activation function. The voxel grid resolution is ${\rm{32}} \times {\rm{32}} \times {\rm{32}}$. The network takes flattened voxel data as input and outputs four-dimensional feature vectors. The latent vector validity check neural network contains three hidden layers with 256 neurons each and uses ReLU activation function. The voxel encoding neural network and the latent vector validity check neural network are trained jointly. For each scene, 2000 latent vector samples are generated, including 1000 valid and 1000 invalid samples. The learning rate is set to 0.0005, the batch size is 256, and the number of training epochs is 150.
	
	\begin{figure*}
		\centering
		\includegraphics[width=0.99\textwidth]{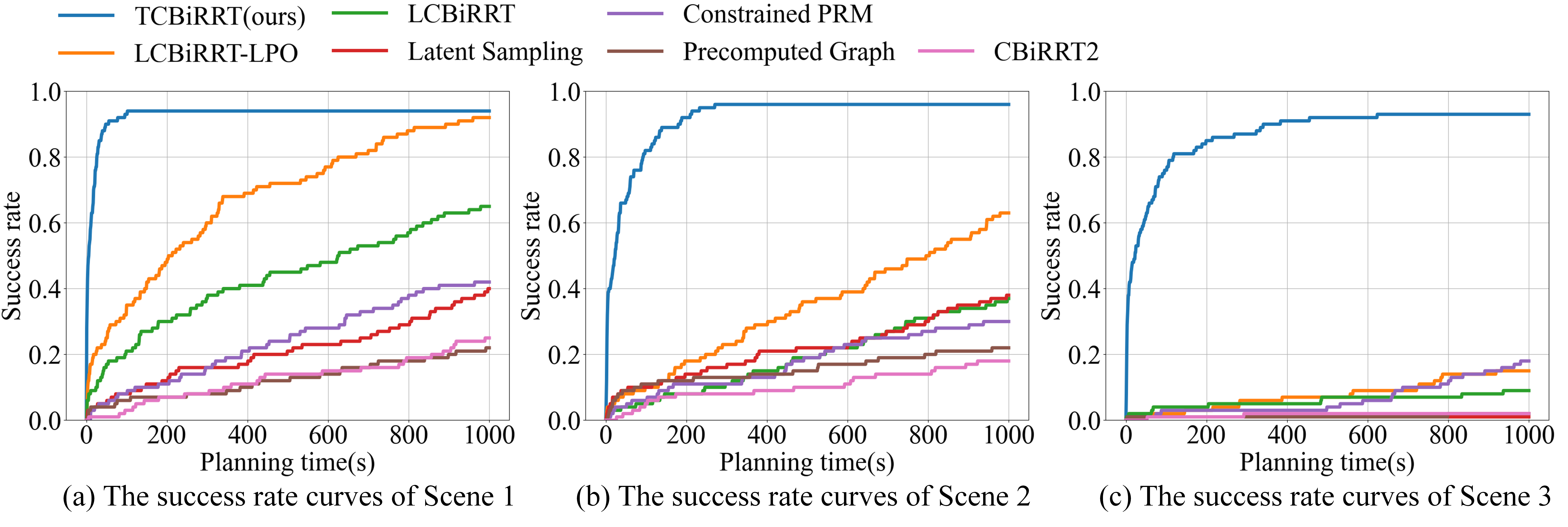}
		\caption{Planning success rate curves in different scenes.}
		\label{fig7}
	\end{figure*}
	
	\item
	\textbf{LCBiRRT-LPO} \cite{c24}\\
	This method extends LCBiRRT by incorporating local path optimization in the latent space. The gradient ascent step size $\gamma$ is set to 0.8, the maximum number of iterations ${N_{\max }}$ is set to 10, and the path optimization interval is 3. The Signed Distance Field (SDF) encoding network contains two hidden layers with 512 neurons each and uses Leaky ReLU activation. The SDF grid size is ${\rm{32}} \times {\rm{32}} \times {\rm{32}}$, and the network outputs 16-dimensional feature vectors. The distance prediction network consists of three hidden layers with 1024 neurons each and uses ReLU activation. For training, 2000 on-manifold configurations are generated per scene, including 1000 collision-free and 1000 in-collision samples. The SDF encoding network networks and the distance prediction network are trained jointly with a learning rate of 0.003, a batch size of 64, and 60 training epochs. Other parameters remain consistent with LCBiRRT.

\end{enumerate}

The TCBiRRT and constrained PRM use the object pose as the sampling variable, while other algorithms explore the full joint space. Therefore, the comparison reflects not only the performance of different search algorithms, but also the effect of different planning-space parameterizations.

\subsection{Evaluation Metrics}

In each scene, 100 pairs of initial and goal object poses are generated. All algorithms are evaluated on the same set of planning tasks to ensure a fair comparison. The performance is assessed in terms of planning success rate and planning time.
The success rate as a function of planning time $t$ is defined as:
\begin{equation}
	p\left( t \right) = \frac{{{n_t}}}{n}
\end{equation}

Where $p(t)$ denotes the success rate under a time limit $t$, $n_t$ is the number of successful planning trials completed within time $t$, and $n$ is the total number of trials ($n = 100$). Let ${T^A}(t) = \{ t_1^A,t_2^A, \cdots ,t_{{n_t}}^A\}$ denote the set of planning times for successful trials of algorithm $A$ within time $t$. The elements in ${T^A}(t)$ are sorted in non-decreasing order, i.e., $t_1^A \le t_2^A \le \cdots \le t_{{n_t}}^A$.

Since different algorithms may yield different numbers of successful trials under the same time limit, a direct comparison of all successful samples may introduce bias. To ensure fairness, only the smallest $n_t^{\min}$ planning times across all algorithms are used for statistical evaluation. The average planning time of algorithm $A$ is computed as:
\begin{equation}
	\bar t_t^A = \frac{1}{{n_t^{\min }}}\sum\limits_{i = 1}^{n_t^{\min }} {t_i^A}
\end{equation}

The corresponding standard deviation is calculated as:
\begin{equation}
	\sigma _t^A = \sqrt {\frac{1}{{n_t^{\min } - 1}}\sum\limits_{i = 1}^{n_t^{\min }} {\left( {t_i^A - {{\bar t_t}^A}} \right)^2} }
\end{equation}

For Scene 1, Scene 2, and Scene 3, the value of $n_t^{\min}$ is set to 40, 30, and 15, respectively. Only algorithms with a number of successful trials greater than or equal to $n_t^{\min}$ are included in the statistical analysis.

\subsection{Simulation Results}


The planning success rate curves are illustrated in Fig. \ref{fig7}, where the success rate $p(t)$ is plotted as a function of the planning time $t$. The success rates at $t = 1000$ s are summarized in Table \ref{table:2}. It can be observed that the proposed TCBiRRT algorithm consistently achieves the highest success rate across all three scenes. In Scene 1, TCBiRRT attains a success rate of 0.94, slightly higher than that of LCBiRRT-LPO. In Scene 2 and Scene 3, the advantage becomes more significant, with success rates of 0.96 and 0.93, respectively, clearly outperforming all baseline methods. In contrast, classical sampling-based constrained planners, such as CBiRRT2, exhibit very low success rates, particularly in Scene 3. This is primarily due to the difficulty of generating valid samples on the constraint manifold in high-dimensional configuration spaces with dense obstacles. Although data-driven approaches, including Latent Sampling and LCBiRRT-LPO, improve planning efficiency, their performance still degrades in complex environments, indicating limited robustness under strong geometric constraints.

The planning time statistics under an equal number of successful trials are presented in Table \ref{table:3}. The results show that TCBiRRT achieves orders-of-magnitude improvements over all baseline methods. Specifically, the average planning times of TCBiRRT are 0.82 s, 2.12 s, and 0.87 s for Scene 1, Scene 2, and Scene 3, respectively, with low standard deviations, demonstrating both efficiency and stability. Compared with the fastest baseline method, LCBiRRT-LPO, the planning speed is improved by factors of 52, 85, and 551 in the three scenes, respectively.

The superior performance of TCBiRRT can be attributed to two main factors. First, the task-space node expansion mechanism enables efficient generation of feasible motion segments without repeated projection onto the constraint manifold, thereby significantly reducing computational overhead. Second, the integration of a bidirectional RRT framework with a regrasp strategy facilitates rapid connection between the two random trees, which reduces the search depth and avoids redundant exploration.

\begin{figure*}[!t]
	\centering
	
	\begin{minipage}{1\linewidth}
		\centering
		\includegraphics[width=0.99\textwidth]{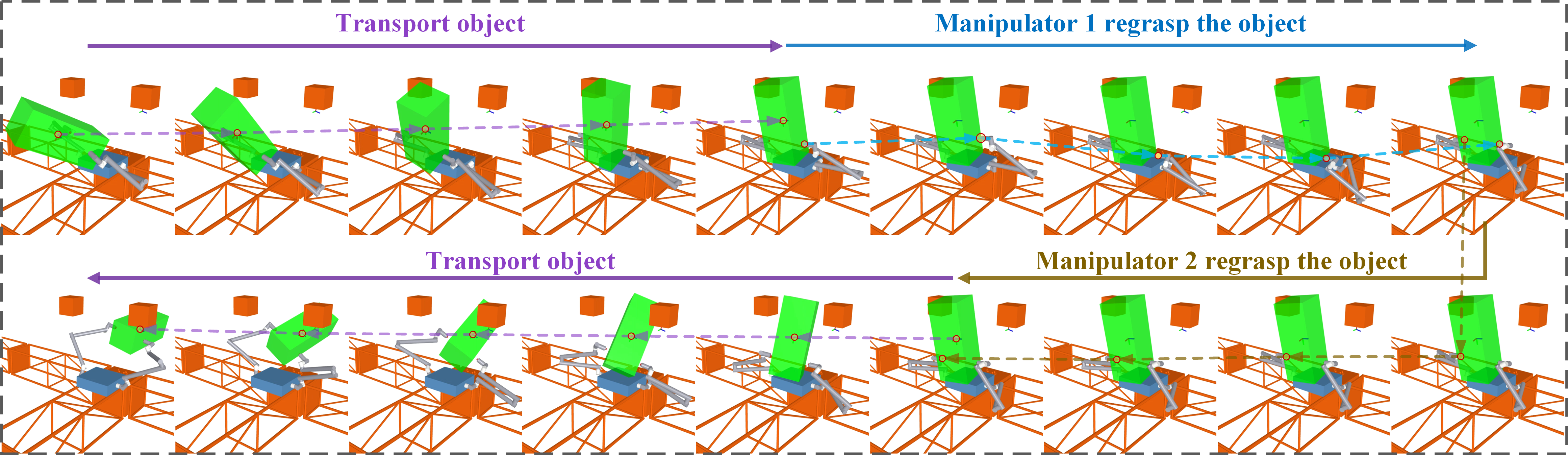}
		\caption{The planned motion path of TCBiRRT algorithm in Scene 1.}
		\label{fig8}
	\end{minipage}
	\hfill
	\begin{minipage}{1\linewidth}
		\centering
		\includegraphics[width=0.99\textwidth]{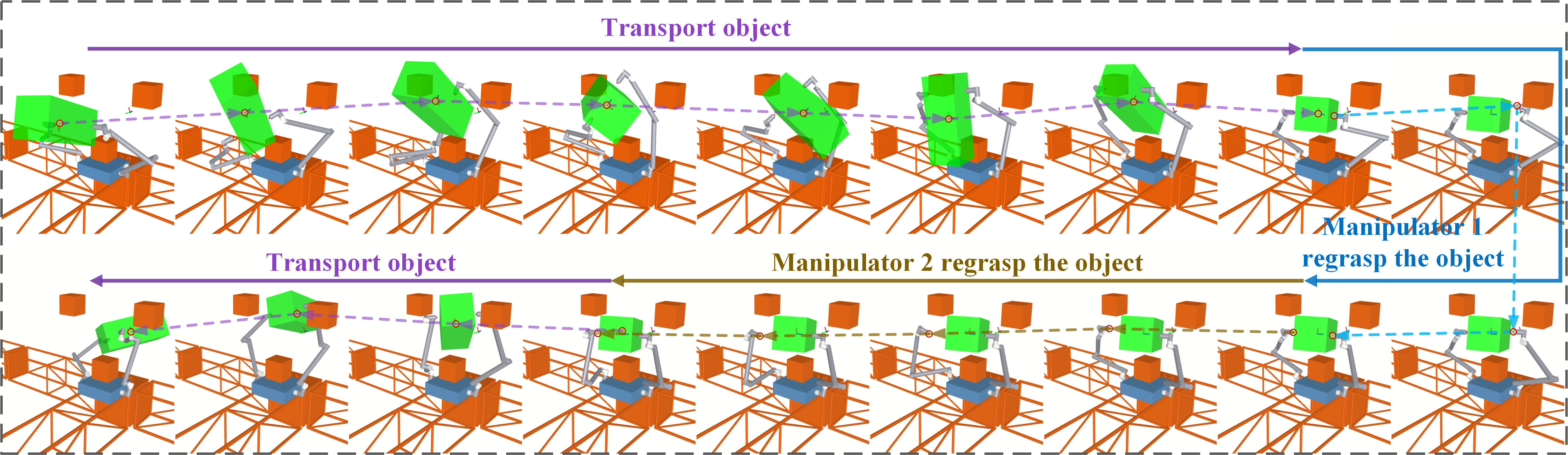}
		\caption{The planned motion path of TCBiRRT algorithm in Scene 2.}
		\label{fig9}
	\end{minipage}
	\hfill
	\begin{minipage}{1\linewidth}
		\centering
		\includegraphics[width=0.99\textwidth]{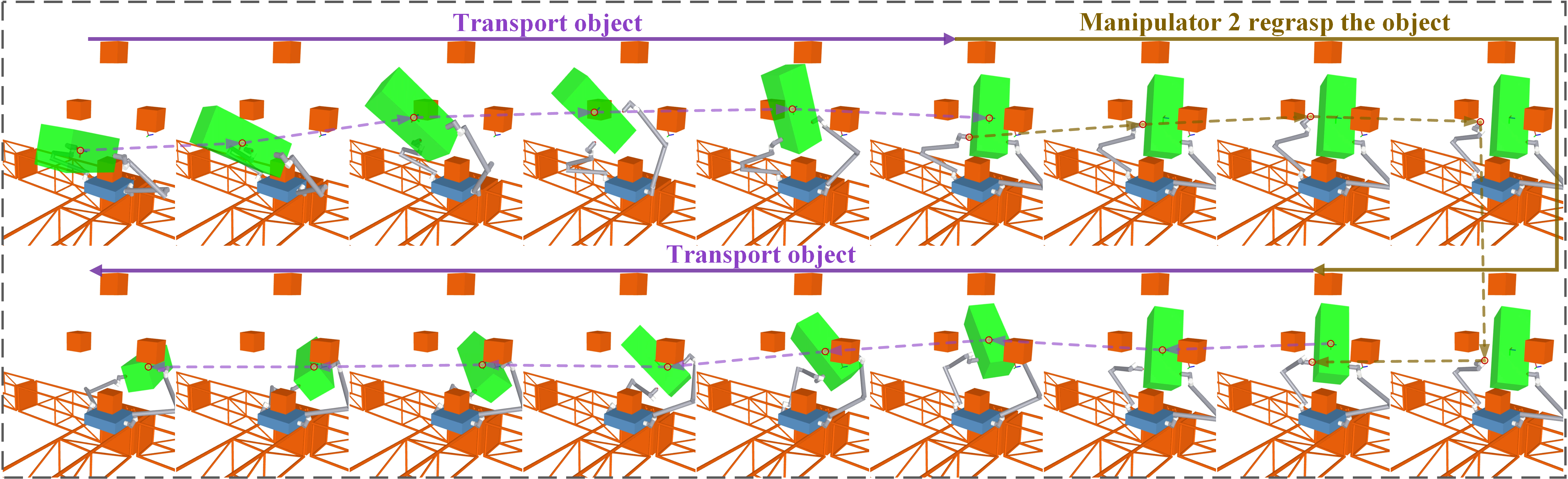}
		\caption{The planned motion path of TCBiRRT algorithm in Scene 3.}
		\label{fig10}
	\end{minipage}
	
\end{figure*}

The motion trajectories of the dual-arm space manipulator in the three scenes are shown in Fig. \ref{fig8} to Fig. \ref{fig10}. When the two random trees meet, the manipulators must transition to new joint configurations sequentially while maintaining the pose of the manipulated object unchanged. This is the regrasp process. The regrasp mechanism reduces planning time at the cost of additional execution steps, which is particularly beneficial in environments with dense obstacles. In practical applications, if the inverse kinematics solutions at the meeting node are sufficiently close, regrasping can be partially or completely avoided. In such cases, only one manipulator or neither needs to adjust its configuration, as illustrated in Fig. \ref{fig10}. 

Table~\ref{table:4} reports the number of successful planning cases in which manipulator 1 performs regrasping, manipulator 2 performs regrasping, and both manipulators perform regrasping. A case is counted if the corresponding manipulator performs regrasp operation. The numbers in parentheses denote the corresponding ratios with respect to the number of successful planning cases of TCBiRRT. The results show that regrasping is frequently used in the proposed TCBiRRT algorithm. This is because the same object pose may correspond to different joint configurations when it is reached from different random trees. Directly connecting these configurations by joint-space interpolation may violate the closed-chain constraint or cause collisions. Therefore, the regrasp mechanism provides a flexible way to connect different inverse-kinematics branches and improves the probability of successfully connecting the two trees.

\begin{table}
	\centering
	\setlength{\abovecaptionskip}{0.cm}
	\caption{The test results of success rate ($t$=1000s)}
	\label{table:2}
	\begin{tabular}{c c c c} 
		\toprule 
		Algorithms & Scene1 & Scene2 & Scene3 \\
		\midrule 
		
		CBiRRT2 \cite{c17}	& 0.25	& 0.18 & 0.02 \\
		Precomputed Graph \cite{c20}	& 0.22	& 0.22 & 0.02 \\
		Constrained PRM  \cite{c15}  & 0.42	& 0.30 & 0.18 \\
		Latent Sampling \cite{c22}	& 0.40	& 0.38 & 0.01 \\
		LCBiRRT	 \cite{c23} & 0.65	& 0.37 & 0.09 \\
		LCBiRRT-LPO  \cite{c24}	& 0.92	& 0.63 & 0.15 \\
		TCBiRRT(ours)	& \textbf{0.94}	& \textbf{0.96} & \textbf{0.93} \\
		
		\bottomrule 
	\end{tabular}
\end{table}

\begin{table}
	\centering
	\setlength{\abovecaptionskip}{0.cm}
	\caption{The test results of planning time (unit: s)}
	\label{table:3}
	\begin{tabular}{c c c c} 
		\toprule 
		Algorithms & \makecell{Scene1\\($n_t^{\min}=40$)} & \makecell{Scene2\\($n_t^{\min}=30$)} & \makecell{Scene3\\($n_t^{\min}=15$)} \\
		\midrule 
		
		CBiRRT2 \cite{c17}	& /	& / & / \\
		\makecell{Precomputed\\Graph \cite{c20}} & /	& / & / \\
		\makecell{Constrained\\RPM  \cite{c15}}  & \makecell{395.19\\(±269.03)}	& \makecell{398.50\\(±287.88)} & \makecell{571.74\\(±288.31)} \\
		\makecell{Latent\\Sampling \cite{c22}}	& \makecell{474.61\\(±346.93)}	& \makecell{301.33\\(±274.02)} & / \\
		LCBiRRT	 \cite{c23}	& \makecell{117.65\\(±108.59)}	& \makecell{386.24\\(±244.36)} & / \\
		\makecell{LCBiRRT-LPO\\ \cite{c24}}	& \makecell{42.89\\(±45.89)}	& \makecell{179.27\\(±130.62)} & \makecell{479.16\\(±301.04)} \\
		TCBiRRT(ours)	& \textbf{\makecell{0.82\\(±0.65)}}	& \textbf{\makecell{2.12\\(±0.94)}} & \textbf{\makecell{0.87\\(±0.38)}} \\
		
		\bottomrule 
	\end{tabular}
\end{table}

\begin{table}
	\centering
	\setlength{\abovecaptionskip}{0.cm}
	\caption{Statistics of regrasp cases performed by the manipulators}
	\label{table:4}
	\begin{tabular}{c c c c}
		\toprule
		Regrasp metric & Scene 1 & Scene 2 & Scene 3 \\
		\midrule
		\makecell{Manipulator 1\\regrasp cases}
		& \makecell{67\\(71.3\%)}
		& \makecell{75\\(78.1\%)}
		& \makecell{72\\(77.4\%)} \\
		
		\makecell{Manipulator 2\\regrasp cases}
		& \makecell{72\\(76.6\%)}
		& \makecell{79\\(82.3\%)}
		& \makecell{83\\(89.2\%)} \\
		
		\makecell{Both manipulators\\regrasp cases}
		& \makecell{63\\(67.0\%)}
		& \makecell{68\\(70.8\%)}
		& \makecell{70\\(75.3\%)} \\
		\bottomrule
	\end{tabular}
	
\end{table}

\subsection{Ablation Study}

To further analyze the contribution of each component, two variants of the proposed method are compared:

\begin{enumerate}
	\def\labelenumi{\arabic{enumi})}
	\item
	\textbf{TCRRT}\\
	Task-space Constrained Rapidly-exploring Random Tree (TCRRT) is a single-tree planner using the same task-space constrained extension operator as TCBiRRT. This variant is used as a single-tree baseline to evaluate the benefit of the complete bidirectional exploration strategy used in TCBiRRT.
	
	\item
	\textbf{TCBiRRT w/o regrasp}\\
	This variant uses the same bidirectional task-space exploration framework as TCBiRRT but removes the regrasp mechanism. When two random trees meet at the same object pose, the corresponding joint configurations $\boldsymbol{q}_a$ and $\boldsymbol{q}_b$ are connected by joint-space interpolation followed by projection onto the constraint manifold, same to the connection strategy used in constrained PRM, as shown in Fig.~\ref{fig4} (b). If all intermediate configurations can be successfully projected onto the constraint manifold and are collision-free, the two trees are considered connected.
\end{enumerate}

\begin{figure*}[t]
	\centering
	\includegraphics[width=0.99\textwidth]{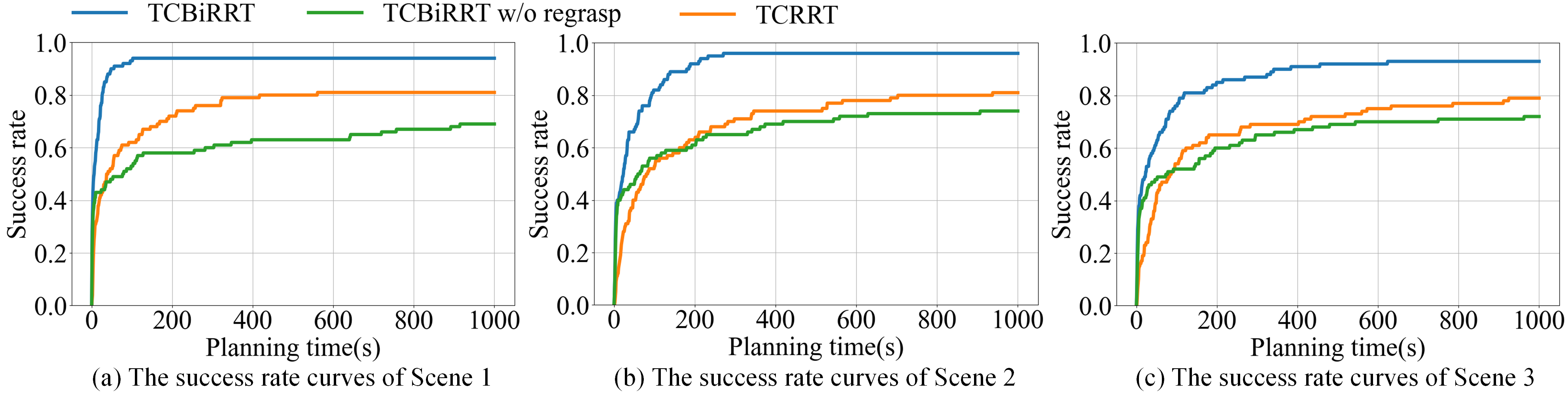}
	\caption{The curves of planning success rates in ablation study.}
	\label{fig11}
\end{figure*}

The planning success rate curves of the ablation study are shown in Fig.~\ref{fig11}. The success rates under a time limit of 1000 s are summarized in Table~\ref{table:5}, and the planning time results are reported in Table~\ref{table:6}. For Scene 1, Scene 2, and Scene 3, the value of $n_t^{\min}$ is set to 65, 70, and 70, respectively. Therefore, the planning times in Table~\ref{table:6} are computed under a different statistical sample size from Table~\ref{table:3}.

The results show that the full TCBiRRT consistently outperforms both variants in all three scenes. Compared with TCRRT, TCBiRRT increases the success rate from 0.81, 0.81, and 0.79 to 0.94, 0.96, and 0.93 in Scenes 1--3, respectively. Meanwhile, the average planning time is reduced from 24.94 s, 68.39 s, and 67.86 s to 3.46 s, 14.03 s, and 16.33 s, corresponding to speedup factors of approximately 7.2, 4.9, and 4.2. This demonstrates that the complete bidirectional connection strategy significantly improves the efficiency of task-space exploration.

The comparison between TCBiRRT and TCBiRRT w/o regrasp further demonstrates the importance of the regrasp mechanism. Without regrasp, the success rates decrease to 0.69, 0.74, and 0.72 in the three scenes, and the average planning times increase to 62.88 s, 62.53 s, and 79.59 s, respectively. In contrast, the full TCBiRRT achieves higher success rates and reduces the planning time by approximately 18.2, 4.5, and 4.9 times. This result indicates that directly connecting two IK branches by joint-space interpolation and projection is unreliable, even when both branches correspond to the same object pose. 

The reason is that the same object pose may correspond to different joint configurations that are far apart from each other. The interpolated joint-space waypoints are numerous, and the joint-space paths obtained through interpolation may significantly deviate from the constraint manifold or collide with obstacles. The regrasp mechanism provides a more flexible way to connect these different IK branches and substantially improves the robustness of bidirectional tree connection.

Overall, the ablation study confirms that the efficiency of TCBiRRT is not solely caused by the reduced dimensionality of the task-space representation. Since all variants use the same six-dimensional planning space, the performance improvement of the full TCBiRRT can be attributed to the combination of task-space constrained extension, bidirectional tree expansion, and regrasp-based connection.

\begin{table}
	\centering
	\setlength{\abovecaptionskip}{0.cm}
	\caption{The success rate results of the ablation study ($t$=1000s)}
	\label{table:5}
	\begin{tabular}{c c c c} 
		\toprule 
		Algorithms & Scene1 & Scene2 & Scene3 \\
		\midrule 
		
		TCRRT	& 0.81	& 0.81 & 0.79 \\
		TCBiRRT w/o regrasp	& 0.69	& 0.74 & 0.72 \\
		TCBiRRT	& \textbf{0.94}	& \textbf{0.96} & \textbf{0.93} \\
		
		\bottomrule 
	\end{tabular}
\end{table}

\begin{table}
	\centering
	\setlength{\abovecaptionskip}{0.cm}
	\caption{The planning time results of the ablation study (unit: s)}
	\label{table:6}
	\begin{tabular}{c c c c} 
		\toprule 
		Algorithms & \makecell{Scene1\\($n_t^{\min}=65$)} & \makecell{Scene2\\($n_t^{\min}=70$)} & \makecell{Scene3\\($n_t^{\min}=70$)} \\
		\midrule 
		
		TCRRT	& \makecell{24.94\\(±29.85)}	& \makecell{68.39\\(±74.23)} & \makecell{67.86\\(±79.03)} \\
		\makecell{TCBiRRT\\ w/o regrasp} & \makecell{62.88\\(±135.88)}	& \makecell{62.53\\(±103.75)} & \makecell{79.59\\(±129.87)} \\
		TCBiRRT	& \textbf{\makecell{3.46\\(±4.24)}}	& \textbf{\makecell{14.03\\(±15.36)}} & \textbf{\makecell{16.33\\(±20.87)}} \\
		
		\bottomrule 
	\end{tabular}
\end{table}

\subsection{Theoretical Discussion}

The proposed TCBiRRT planner generates feasible motions by combining task-space exploration with path inverse kinematics. For a given object pose $\boldsymbol{\xi}_o$, the desired pose of each end-effector is uniquely determined by the fixed grasp transformation between the object and the end-effector. Therefore, if the inverse kinematics solver successfully finds the joint angles of both manipulators for the desired end-effector poses, the relative transformation between the two end-effectors and the manipulated object remains unchanged. In this sense, the closed-chain constraint is satisfied up to the numerical tolerances of the inverse kinematics solver and the task-space interpolation step.

Let $\varepsilon_p$ and $\varepsilon_o$ denote the position and orientation error thresholds used in the inverse kinematics solver. For each waypoint of the interpolated object path, the generated joint configuration is accepted only when the end-effector position and orientation errors of both manipulators are smaller than $\varepsilon_p$ and $\varepsilon_o$, respectively. Therefore, the closed-chain constraint error is bounded by the inverse kinematics accuracy. In addition, all interpolated configurations are checked for collision, and an edge is added to the random tree only when the complete joint sequence is collision-free.

The success of the PathInverseKinematics procedure is local and depends on several conditions. When the task-space interpolation step is sufficiently small, the desired end-effector poses remain inside the reachable workspace, the manipulator Jacobian is nonsingular or well conditioned, and the path does not encounter joint limits or collisions, the solution at the previous waypoint provides an effective seed for the next waypoint. Under these conditions, the iterative inverse kinematics solver can continuously track the task-space path and generate a continuous joint-space trajectory. However, if the object path passes through a singular region, an unreachable pose, or a narrow collision region, PathInverseKinematics may fail and the corresponding tree extension is rejected.

The proposed method inherits the exploratory nature of sampling-based planning, but it is not claimed to be globally probabilistically complete on the full closed-chain constraint manifold. This is because the planner samples in the object task space and recovers joint configurations using a deterministic inverse kinematics procedure. As a result, the reachable portion of the constraint manifold depends on the selected IK branch, the seed configuration, and the feasibility of regrasp connections. Nevertheless, in the considered tightly coupled transportation task, the object pose provides a compact and physically meaningful parameterization of the dominant motion, which significantly improves practical planning efficiency.

\section{Conclusion}
This paper presents TCBiRRT, a task-space constrained bidirectional RRT algorithm for tightly coupled dual-arm space manipulators under closed-chain constraints. By transforming the planning problem from the high-dimensional configuration space to a lower-dimensional task space, the proposed method significantly improves the efficiency of node expansion. The integration of the bidirectional RRT framework and the regrasp mechanism further enhances the connectivity of the search process and enables efficient handling of multiple inverse kinematics solutions.

Simulation results in representative on-orbit assembly scenarios demonstrate that the proposed method achieves substantially higher success rates and significantly lower planning times compared to existing approaches. The performance gains are particularly evident in cluttered environments, where conventional methods suffer from low sampling efficiency or limited robustness.

The proposed method should be interpreted as a high-level kinematic planner for the transportation phase of on-orbit assembly. It aims to rapidly generate an initial feasible path that satisfies collision avoidance and closed-chain kinematic constraints. However, joint acceleration, torque limits, microgravity dynamics, and flexible structural effects are not explicitly considered in the current formulation. Therefore, the planned path may require further post-processing, such as shortcutting, smoothing, time parameterization, dynamic feasibility checking, and robust tracking control. These issues are beyond the scope of the present work but are important directions for future research. Future work will extend the proposed framework toward kinodynamic planning by incorporating dynamic constraints into the local planner, adding uncertainty margins to collision checking, and using the generated path as an initial solution for trajectory optimization.

\section{Appendix}
For a given $\boldsymbol{R} \in SO(3)$, there is always a unit rotation axis ${\boldsymbol{\hat \omega }} \in {\mathbb{R}^3} \left( \left\| {{{\boldsymbol{\hat \omega }}}} \right\|{\rm{ = }}1 \right) $ and a rotation angle $\theta$ such that $\boldsymbol{R} = {e^{[\boldsymbol{\hat \omega} ]\theta }}$. Vector $\boldsymbol{\hat \omega } \theta \in {\mathbb{R}^3}$ is the rotation exponential coordinate of $\boldsymbol{R}$, which can be obtained by separately  calculating the $\theta$ and the $\boldsymbol{\hat \omega }$. The formula for calculating the $\theta$ is as follows:
\begin{equation}
	\theta ={\rm{arccos}}\left( {\frac{{{\rm{tr}}\left( {{\boldsymbol{R}}} \right) - 1}}{2}} \right)
\end{equation}

Where ${\rm{tr}}( \cdot )$ represents the calculation of the trace of a matrix. For the unit rotation axis ${\boldsymbol{\hat \omega }}$, if $\theta = 0$: ${\boldsymbol{\hat \omega }} = {\boldsymbol{0}_3}$. 

If $\theta = \pi$:
\begin{equation}
	\boldsymbol{\hat \omega }=\frac{1}{{\sqrt {2\left( {1 + {r_{11}}} \right)} }}\left[ {\begin{array}{*{20}{c}}
			{{1+r_{11}}}\\
			{{r_{21}}}\\
			{{r_{31}}}
	\end{array}} \right]
\end{equation}

Where $r_{ab}$ represents the element in the a-th row and b-th column of the rotation matrix $\boldsymbol{R}$.

If $\theta \neq \pi$ and $\theta \neq 0$:
\begin{equation}
	\boldsymbol{\hat \omega } = \frac{1}{{2\sin \theta }}\left[ {\begin{array}{*{20}{c}}
			{{r_{32}} - {r_{23}}}\\
			{{r_{13}} - {r_{31}}}\\
			{{r_{21}} - {r_{12}}}
	\end{array}} \right]
\end{equation}

\end{document}